%% file: main.tex
\documentclass[conference]{IEEEtran}
\include{_header}

\newcommand{\tsea}{{T-SEA}\xspace}
\newcommand{\ours}{{\sf P$^3$A}\xspace}

\title{
Towards Powerful and Practical Patch Attacks for 2D Object Detection in Autonomous Driving}
\author{
Yuxin Cao\textsuperscript{1,2} \quad
Yedi Zhang\textsuperscript{2}\thanks{Corresponding author: \href{mailto:yd.zhang@nus.edu.sg}{yd.zhang@nus.edu.sg}} \quad
Wentao He\textsuperscript{3} \quad
Yifan Liao\textsuperscript{2} \quad
Yan Xiao\textsuperscript{4} \quad
\\
Chang Li\textsuperscript{1} \quad
Zhiyong Huang\textsuperscript{2} \quad
Jin Song Dong\textsuperscript{2}
\\
\\
\textsuperscript{1}\textit{State Key Laboratory of Intelligent Vehicle Safety Technology, Changan Automobile, Chongqing, China} \\
\textsuperscript{2}\textit{National University of Singapore, Singapore} \\
\textsuperscript{3}\textit{Ningbo University, Ningbo, China} \\
\textsuperscript{4}\textit{Sun Yat-Sen University, Guangzhou, China}
}
\date{}

\begin{document}

\maketitle

\begin{abstract}
Despite advances, learning-based autonomous driving systems remain critically vulnerable to adversarial patches, posing serious safety and security risks in their real-world deployment. Black-box attacks, notable for their high attack success rate without model knowledge, are especially concerning, with their transferability extensively studied to reduce computational costs compared to query-based attack methods. Previous transferability-based black-box attacks typically adopt mean Average Precision (mAP) as the evaluation metric and design training loss accordingly. However, due to the presence of multiple detected bounding boxes and the relatively lenient Intersection over Union (IoU) thresholds, the attack effectiveness of these approaches is often overestimated, resulting in reduced success rates in practical attacking scenarios. Furthermore, patches trained on low-resolution data often fail to maintain effectiveness on high-resolution images, limiting their transferability to high-resolution autonomous driving datasets. To fill this gap, we propose \ours, a \textbf{P}owerful and \textbf{P}ractical \textbf{P}atch \textbf{A}ttack framework for 2D object detection in autonomous driving, specifically optimized for high-resolution datasets. First, based on IoU, we introduce a novel evaluation metric, Practical Attack Success Rate (PASR), to more accurately quantify adversarial patch attack effectiveness with greater relevance for pedestrian safety in autonomous driving. Second, we present a tailored loss function, Localization-Confidence Suppression Loss (LCSL), to improve attack transferability under PASR. Finally, to maintain the transferability for high-resolution datasets, we further incorporate the Probabilistic Scale-Preserving Padding (PSPP) into the patch attack pipeline as a data preprocessing step. Extensive experiments show that \ours outperforms state-of-the-art attacks on unseen models and unseen high-resolution datasets, both under the proposed practical IoU-based evaluation metric and the previous mAP-based metrics.
\end{abstract}

\section{Introduction}\label{sec:intro}
Learning-driven 2D object detection has achieved human-level performance in applications such as autonomous driving~\cite{feng2021review,mao20233d}, but remains vulnerable to adversarial perturbations~\cite{szegedy2014intriguing}, which threaten the reliability and safety of real-world deployment~\cite{chi2024adversarial,wang2023does}. Among various attack methods, adversarial patch attacks have attracted particular attention due to their practicality in physical settings~\cite{wei2022adversarial}. Initially targeting traffic sign recognition, these attacks have evolved to compromise pedestrian detection~\cite{huang2020universal,xu2020advtshirt}, posing serious risks to autonomous vehicles. Unlike query-based perturbations~\cite{ilyas2018black}, patches are often trained on an ensemble of multiple surrogate models to enhance transferability across models, datasets, and environments~\cite{hu2021naturalistic}, shifting the burden from test-time queries to offline training. Recent works such as \tsea~\cite{huang2023tsea} further improve transferability by employing self-ensembling strategies, demonstrating effective attacks using only one single surrogate model.

\begin{figure}[t]
  \centering
  \includegraphics[width=1\linewidth]{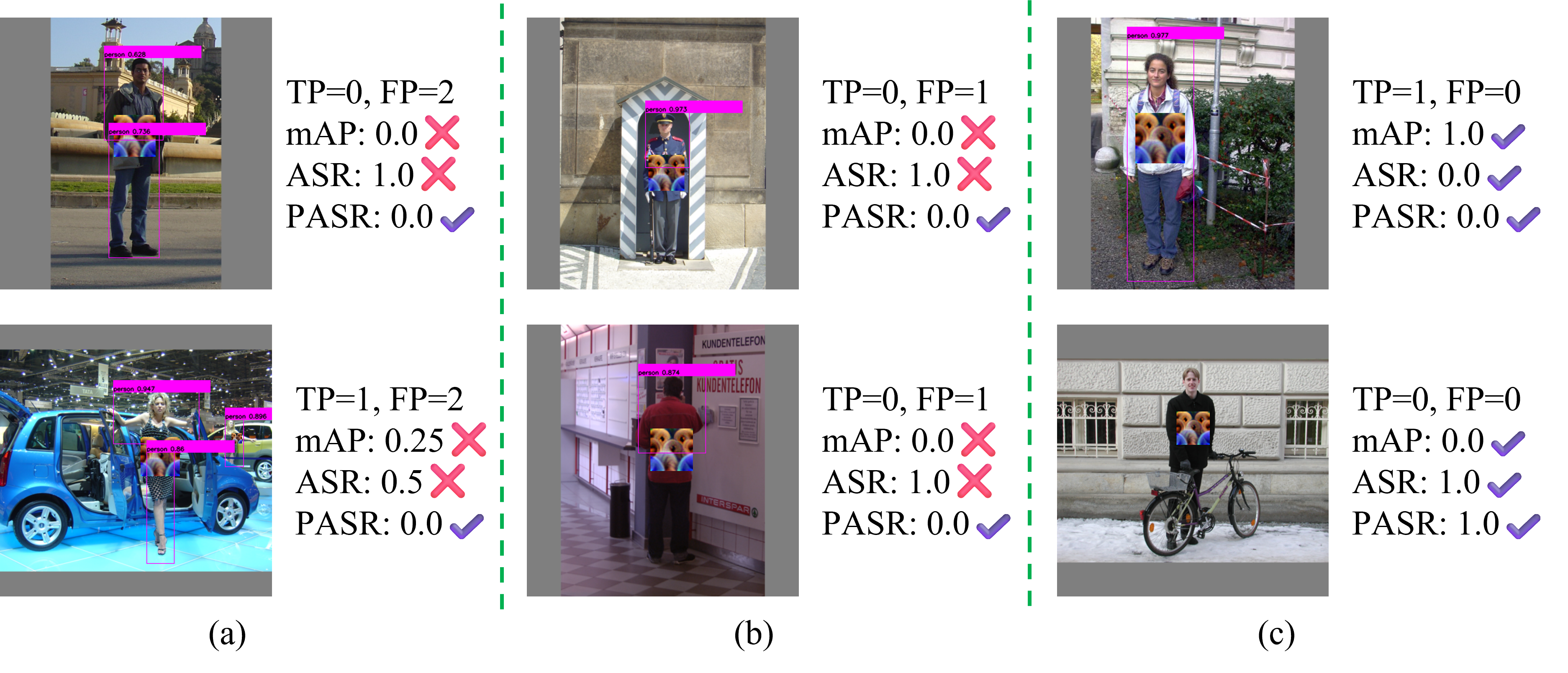}
  \caption{Examples of the overestimation issues inherent in mAP-based attack metrics (mAP and ASR). (a) Multiple detection boxes. (b) Below-threshold IoU matching. (c) Attack failure and success. Despite the mAP dropping to nearly zero across all images in (a) and (b), no person is completely undetected. However, our proposed PASR can always reflect whether the person is hidden. Patch is from \tsea.
  }
  \label{fig:overestimation}
\end{figure}

Most existing research on transferable patch attacks relies heavily on mean Average Precision (mAP) as the primary evaluation metric, with corresponding loss functions and optimization strategies specifically designed to minimize mAP. However, 
as shown in Figure~\ref{fig:overestimation}, the mAP value typically drops after a patch attack due to two main factors: i) Multiple overlapping detection boxes are predicted for a single pedestrian, and ii) the Intersection over Union (IoU) value between the predicted and the ground truth (GT) bounding box falls below the IoU threshold. Specifically, the presence of multiple overlapping boxes leads to an increase in false positives (FP), thereby lowering the precision component of mAP. Moreover, if all detected boxes have an IoU below the threshold, the corresponding GT cannot be matched to any prediction, resulting in a loss of true positive (TP) and further degrading both precision and recall, thus contributing to the decline in mAP. Since such cases in autonomous driving do not prevent the vehicle from stopping in front of pedestrians, they do not pose a real safety hazard~\cite{wei2024cap}.
Thus, a more practical metric is urgently needed --- one that can effectively reflect whether a person is directly ignored by the detector.
Furthermore, existing patches are trained on low-resolution (LR) data ($\approx$600 pixels), and perform poorly on high-resolution (HR) autonomous driving data ($\ge$1200 pixels), where pedestrians occupy smaller regions and patches struggle to dominate detection.

\noindent
{\bf Contributions.} In this work, we introduce a Powerful and Practical Patch Attack framework, \ours, to improve the practical attack transferability of adversarial patches towards HR autonomous driving datasets. To achieve this, we first propose Practical Attack Success Rate (PASR), a new evaluation metric that \textit{more accurately} reflects real-world risks by considering an attack successful only when no detection box overlaps with the GT pedestrian, as such situation would lead a direct collision with the pedestrian in real-world scenarios. 
Inspired by the new metric and the IoU loss used in the model training phase, we propose a Localization-Confidence Suppression Loss (LCSL), which simultaneously leverages IoU and confidence scores to facilitate attacks and improve transferability. Note that almost all existing works rely solely on confidence scores (objectness \& classification score) while overlooking the crucial role of IoU in enhancing attack effectiveness. 
Finally, we propose the Probabilistic Scale-Preserving Padding (PSPP) to maintain the absolute size of person in HR data, which can enhance the transferability of the trained patch to autonomous driving datasets. 
Through comprehensive experiments, we show that \ours outperforms state-of-the-art (SOTA) attacks in terms of PASR, as well as mAP-based metrics, on 11 mainstream object detectors and nine widely used datasets. 
Specifically, \ours can achieve an PASR up to 52\% for model transferability and 54\% for data transferability.

Our contribution can be summarized as follows.
\begin{itemize}
    \item We summarize the overestimation issues of existing mAP-based metrics and propose a novel metric, PASR, which accurately reflects the practical attack performance of adversarial patches.
    \item Motivated by the IoU loss used in object detector training, we propose a Localization-Confidence Suppression Loss (LCSL), which fully utilizes both information to enhance the effectiveness and transferability of the trained patch.
    \item We propose a novel black-box patch attack framework, \ours, that addresses the limited transferability of existing patches to HR autonomous driving data. 
    \item We show that \ours outperforms SOTA attacks on 11 detectors and nine datasets (including seven HR datasets), demonstrating stronger transferability.
\end{itemize}

\section{Related Work}\label{sec:related}
\noindent\textbf{Black-box Adversarial Attacks.}
Black-box adversarial attacks have garnered increasing attention in recent years due to their practical applicability in real-world scenarios, particularly when targeting commercial models.
These attacks are typically categorized into query-based~\cite{ilyas2018black,liang2022parallel} and transfer-based attacks~\cite{cai2022zero,wang2025transferable}, depending on the level of access to the target model. Query-based attacks require extensive model access, which usually leads to high costs and easily raises alarms. In contrast, transfer-based attacks generate adversarial examples by training on local surrogate models and do not rely on repeated queries. Since these attacks can be executed in real time, they are more practical and scalable, especially in restricted or high-risk environments.

\noindent\textbf{Transfer-based Patch Attacks in Object Detection.}
Adversarial patches have been widely used to attack object detectors, particularly in person detection tasks. Due to the need for real-time responsiveness and strong transferability, transfer-based attacks have emerged as a mainstream attack approach. Existing work primarily focuses on either improving the stealthiness of adversarial patches~\cite{hu2021naturalistic,tan2021lap,zhou2023mvpatch} or enhancing the attack effectiveness~\cite{thys2019advpatch,cheng2024depatch}. To ensure the cross-model transferability, the patch can be trained locally using an ensemble of diverse detectors~\cite{hu2021naturalistic,zhou2023mvpatch}, promoting generalization to unseen models. \tsea~\cite{huang2023tsea} proposes a self-ensemble framework that achieves state-of-the-art cross-model transferability while requiring only a single surrogate model for patch training. On the other hand, existing transfer-based attacks typically train adversarial patches solely on LR datasets (\eg INRIA~\cite{dalal2005inria}),  usually exhibiting poor attack performance when applied to HR autonomous driving scenes. Additionally, while the commonly used metric mAP is effective for general object detection, it becomes less reliable in scenarios involving multiple overlapping detections and below-threshold IoU matching. 
In this paper, we propose a novel practical metric for the community to more accurately and reliably evaluate practical patch attacks, particularly in autonomous driving contexts. To improve attack transferability under this metric, we incorporate IoU-based loss into the attack process and focus on addressing the HR dataset challenge.

\section{Preliminary}
\noindent\textbf{Threat Model.}
\ours operates under two realistic constraints. First, consistent with \tsea, the attacker can access only a single surrogate detector during training, which reduces computational cost and aligns with real-world limitations where identifying and training detectors is resource-intensive. Second, training is limited to LR data (\eg INRIA), as HR autonomous driving datasets (\eg $\ge$1K) are costly to collect and rarely released publicly. Moreover, HR datasets require higher training costs and may lack generalization due to domain specificity. Given these constraints, our key research question is: how to improve patch transferability to black-box detectors and unseen HR autonomous driving data using only one surrogate and one LR dataset?

\noindent\textbf{Problem Formulation.}
Given an input image, an object detector produces $N$ candidate predictions $\bar {\mathbf{O}} = \left\{ \left({{{{\bf{\bar b}}}_i},{{\bar o}_i},{{{\bf{\bar s}}}_i}} \right) \right\}_{i = 1}^N$, 
where ${{{\bf{\bar b}}}_i}$ denotes the bounding box coordinates, ${\bar o}_i$ denotes the objectness score, and ${{{\bf{\bar s}}}_i} \in \mathcal{R}^C$ denotes the classification score over $C$ classes. 
Predictions with confidence scores below a confidence threshold are first discarded, and then Non-Maximum Suppression (NMS) is applied to the remaining boxes, suppressing those with IoU above a given IoU threshold to eliminate duplicates. 
The final prediction set is $\mathbf{O} = \left\{ \left({{\mathbf{b}_j},{p_j},{c_j}} \right) \right\}_{j = 1}^M$, 
where $\mathbf{b}_j$ denotes the retained bounding box, ${c_j}$ denotes the predicted class, and $p_j$ is the final confidence score (product of objectness score ${\bar o}_j$ and classification score ${\bar s}_j^{c_j}$ for class $c_j$).

In object detector attacks, the prediction set $\mathbf{O}(\mathbf{x})$ for the clean image $\mathbf{x}$ is regarded as the ground truth (GT)~\cite{hu2021naturalistic,huang2023tsea}. The attacker then generates a patch ${\mathbf{\xi }}$ applied within $\mathbf{b}_j$ through a mask $\mathbf{M}$, producing $\mathbf{x}_{adv} = \mathbf{x} \odot (1-\mathbf{M}) + \mathbf{\xi } \odot \mathbf{M}$. Hence, the attacker's goal is to find a $\xi$ to cause $\mathbf{O}(\mathbf{x}_{adv}) \ne \mathbf{O}(\mathbf{x})$. Since object detection involves both regression and classification, the traditional metric mAP considers an attack successful if 
\begin{equation}\label{equ:attack}
{\rm{IoU}}( {{{\bf{b}}_j}( {{\mathbf{x}_{adv}}} ),{{\bf{b}}_j}( \mathbf{x} )} ) < 0.5 \quad {\rm{or}} \quad {c_j}( {{\mathbf{x}_{adv}}} ) \ne {c_j}( \mathbf{x} ),
\end{equation}
where 0.5 is often used as the IoU threshold in object detection. However, we will show that such successful attacks may not prevent the autonomous vehicles from stopping.

\section{Analyses for the Existing Techniques}

\subsection{Attack Metrics}\label{sec:metrics}
Table~\ref{tab:metrics_losses_comparison} summarizes the existing mainstream methods for generating transferable adversarial patches. Almost all methods use mAP as the evaluation metric for digital attacks, and several methods additionally use the Attack Success Rate (ASR) to assess the effectiveness of physical attacks only. 
However, we notice that mAP and ASR both suffer from overestimation issues of multiple detection boxes and below-threshold IoU matching. Therefore, both metrics do not reflect the true attack impact.

\noindent\textbf{Mean Average Precision.}
The definition of mAP is: 
\begin{equation}
\begin{array}{l}
{\rm{P}} = \frac{{{\rm{TP}}}}{{{\rm{TP + FP}}}},{\rm{R}} = \frac{{{\rm{TP}}}}{{{\rm{TP + FN}}}},\\
{\rm{mAP}} = \frac{1}{{n - 1}}\sum\limits_{i = 1}^{n - 1} {\left( {{{\rm{R}}_{i + 1}} - {{\rm{R}}_i}} \right){{\rm{P}}_{i + 1}}} ,
\end{array}
\end{equation}
where TP, FP, TN, and FN denote respectively true positive, false positive, true negative, and false negative samples, P denotes precision and R denotes recall. mAP is computed as the area under the P-R curve. Although widely used, IoU faces the following two overestimation problems.

\begin{table}[t]  
\centering
\caption{Metrics and losses used in transferable patch attacks. obj: objectness score, cls: classification score.
}
\label{tab:metrics_losses_comparison}
\begin{tabular}{lll}
\toprule
Research & Metric & Adversarial Loss \\
\midrule
AdvPatch & mAP & obj/cls/obj$\times$cls \\
NAP & mAP & obj$\times$cls  \\
LAP & mAP & obj \\
\tsea & mAP & obj \\
AdvART & mAP & obj$\times$cls \\
AdvTexture & mAP, ASR & obj \\
MVPatch & mAP, ASR & cls \\
DePatch & mAP, ASR & obj, affiliated by IoU \\
FDA & ASR & cls+IoU \\
\midrule
\ours & PASR, mAP, ASR & obj$\times$cls$\times$IoU \\
\bottomrule
\end{tabular}
\end{table}

1) \textbf{Multiple Detection Boxes:}
As illustrated in Figure~\ref{fig:overestimation}(a), when a single object is detected by multiple bounding boxes with overlaps, the detector fails to match any of them to the GT since there is no IoU greater than 0.5 between any predicted box and the GT box. It seems that the original prediction box is fragmented into several bounding boxes due to the adversarial patch, resulting in multiple FPs and no TP. The increase in FP leads to a decrease in precision, and the loss of TP reduces both precision and recall, both contributing to a drop in mAP.
As a result, these situations exaggerate the effectiveness of attacks and fail to accurately reflect their true impact on detection performance.

2) \textbf{Below-Threshold IoU Matching:}
As shown in Figure~\ref{fig:overestimation}(b), when the detector predicts only one bounding box for an object but its IoU with the GT box is below threshold (typically 0.5), the prediction is treated as a FP rather than a TP, which, similarly, leads to a decrease in mAP. However, it cannot be neglected that these smaller boxes still capture meaningful semantic information of the person (\eg the upper part of the body). From the perspective of an autonomous vehicle, such detections would still trigger appropriate responses, such as slowing down or stopping to avoid the ``smaller'' pedestrian. Therefore, although mAP decreases, the functionality of the object detector remains unaffected, and the attack does not truly endanger safety.

To sum up, these two overestimation issues illustrate a key limitation of mAP: it overstates the impact of patch attacks by penalizing non-critical bounding boxes (decreasing TP and increasing FP). Hence, relying solely on mAP for safety assessment in both digital and physical scenarios can lead to misleading conclusions about the attack performance.

\noindent\textbf{Attack Success Rate.}
In a limited number of studies, ASR is used to evaluate physical attacks, particularly in real-world scenarios. The definition of ASR is as follows.
\begin{equation}
{\rm{ASR}} = 1 - \frac{{{\rm{TP}}}}{{{\rm{TP}'}}},
\end{equation}
where ${\rm{TP}'}$ denotes the number of GT samples. ASR measures the proportion of the objects for which the detector fails to produce a valid detection, \ie Equation~\ref{equ:attack}. 
Although ASR focuses solely on TPs and does not penalize redundant or imprecise bounding boxes, it still adopts the same definition of TPs as mAP. Therefore, ASR can be regarded as a mAP-based metric. As shown in Figure~\ref{fig:overestimation}, ASR also suffers from the above two overestimation issues: when a person is detected by one or more bounding boxes, all of which have an IoU less than the threshold with the GT box, ASR incorrectly treats the case as a successful attack, since none of the predicted boxes is counted as a TP. Yet, such detections may still retain sufficient semantic cues and cannot prevent autonomous vehicles from slowing down or stopping, and therefore, do not potentially pose a safety threat. We argue that such cases should not be regarded as a successful attack.

\subsection{Adversarial Losses}\label{sec:adv_loss}
\noindent\textbf{Score-based Loss.}
Recall that the detector outputs two kinds of scores: objectness score ${\bar o}$ and classification score $\bf{\bar s}$. Lowering either score can increase the likelihood of the bounding boxes being discarded during the NMS stage, potentially resulting in the removal of the original bounding box. Therefore, to effectively mislead the detector, AdvPatch~\cite{thys2019advpatch} first proposed three score-based adversarial losses, with minimizing the maximum objectness score performing the best. This strategy has since been widely adopted, as shown in Table~\ref{tab:metrics_losses_comparison}.

\noindent\textbf{IoU-based Loss.}
In recent years, IoU has started to receive limited attention in a few studies, and has been incorporated into the design of adversarial loss, despite not being explored in sufficient detail. Depatch~\cite{cheng2024depatch} selects the best bounding box using a weighted sum of objectness score and IoU, but only optimizes the objectness score corresponding to that box, leaving IoU unused as a direct optimization objective. 
To ensure patch effectiveness across varying distances, FDA~\cite{cheng2024fda} optimizes the average of the combination of classification score and IoU. Although averaging may align well with their use of Expectation Over Transformation, it introduces significantly higher computational overhead, compared to optimizing the maximum. 
In the field of optical remote sensing, PA-ORSI~\cite{sun2023threatening} designs adversarial patches based on the sum of the objectness score and the maximum IoU to avoid gradient inundation. However, the two terms are treated independently, which fails to capture their joint effect. For example, the bounding box with the highest IoU may still have a low objectness score, which weakens the attack effectiveness. 
Overall, existing approaches fail to fully exploit the joint impact of IoU and confidence scores, which is critical for improving attack transferability.

\section{\ours Design}
\subsection{Practical Attack Success Rate}
According to the analyses above, we conclude that relying on Equation~\ref{equ:attack} to measure attack performance cannot truly reflect the actual risk in real-world scenarios. To address this limitation, we propose Practical Attack Success Rate (PASR), a new metric that more effectively captures the practical impact of adversarial patch attacks. 

Formally, given the GT boxes of an image $\mathcal{G} = \left\{ {\bf{b}}_1^{\text{GT}}, {\bf{b}}_2^{\text{GT}},... \right\}$ and the predicted bounding boxes of a detector $\mathcal{D} = \left\{ {{\bf{b}}_1^d,{\bf{b}}_2^d,...} \right\}$, we first define that a GT box ${{\bf{b}}_i^{\text{GT}}}$ is successfully attacked if \textbf{none} of the predicted boxes has any intersection with it. The object-level attack success indicator function can be expressed as:
\begin{equation}\label{equ:object_level}
A\left( {{\bf{b}}_i^{\text{GT}}} \right) = \left\{ \begin{array}{l}
1,{\rm{if}}\ \forall {\bf{b}}_j^d \in {\mathcal D},{\rm{IoU}}\left( {{\bf{b}}_i^{\text{GT}},{\bf{b}}_j^d} \right) = 0,\\
0,\rm{otherwise}.
\end{array} \right.
\end{equation}

This strict definition ensures that even minor overlaps between the predicted boxes and the GT box will \textit{disqualify} the attack as successful, aligning with the intuition that if the detector still perceives any part of the object, the attack fails from a safety perspective. 
Then we escalate the evaluation from the object level (\eg mAP, ASR) to the image level, as in autonomous driving, the existence of even one undetected pedestrian is sufficient to raise safety concerns. We define that an image is considered successfully attacked if \textbf{any} of its GT box is successfully attacked: 
\begin{equation}\label{equ:image_level}
\tilde A\left( \mathbf{x} \right) = \left\{ \begin{array}{l}
1,{\rm{if}}\ \exists {\bf{b}}_i^{\text{GT}} \in {\mathcal G},A\left( {{\bf{b}}_i^{\text{GT}}} \right) = 1,\\
0,\rm{otherwise}.
\end{array} \right.
\end{equation}

Finally, given a dataset of images, we first filter out images where the detector cannot output any boxes for persons. Suppose that this yields a set of $T$ remaining images. 
The adversarial patch is overlaid onto each of the person boxes to generate the adversarial images $\left\{ {{\mathbf{x}}_{adv}^1,{\mathbf{x}}_{adv}^2,...,{\mathbf{x}}_{adv}^T} \right\}$. Then, we compute PASR as the proportion of these adversarial images that are successfully attacked: 
\begin{equation}
{\rm{PASR}} = \frac{1}{T}\sum\nolimits_{t = 1}^T {\tilde A\left( {\mathbf{x}_{adv}^t} \right)} .
\end{equation}

In summary, PASR operates at the image level and provides a safety-oriented evaluation by requiring the most stringent criterion for attack success. This is particularly suitable for physical attack scenarios, where the attack performance is often evaluated by how many frames in a video sequence the attack is effective, rather than the object-level accuracy of each detection. See Appendix for additional justification and regulatory context.

\subsection{Localization-Confidence Suppression Loss}
Using existing score-based strategies alone can only reduce the confidence of detection boxes, making the detector less likely to output a high confidence score for boxes that have a large IoU. As a result, boxes with lower IoU and moderate confidence may become the final outputs after NMS. This leads to a certain drop in mAP, but cannot reduce PASR.

We conducted a pilot experiment using IoU alone as the adversarial loss, and the results show that it fails to achieve successful attacks, even on a single image. This is because detection results are jointly determined by both the bounding box location and its associated confidence score. Although optimizing only IoU can reduce it to around 0.6, it does not sufficiently reduce the confidence score, and thus the object remains detectable. Conversely, optimizing only the confidence score can lead to a certain degree of attack success, since when the confidence score of a high-IoU box drops below the confidence threshold, the detection box still exists but will not be output. Actually, existing attacks exploit the imbalance between the classification and localization components of object detection and result in attacks that are effective under mAP-based metrics.

\begin{figure*}[!t]
    \centering
    \includegraphics[width=0.8\linewidth]{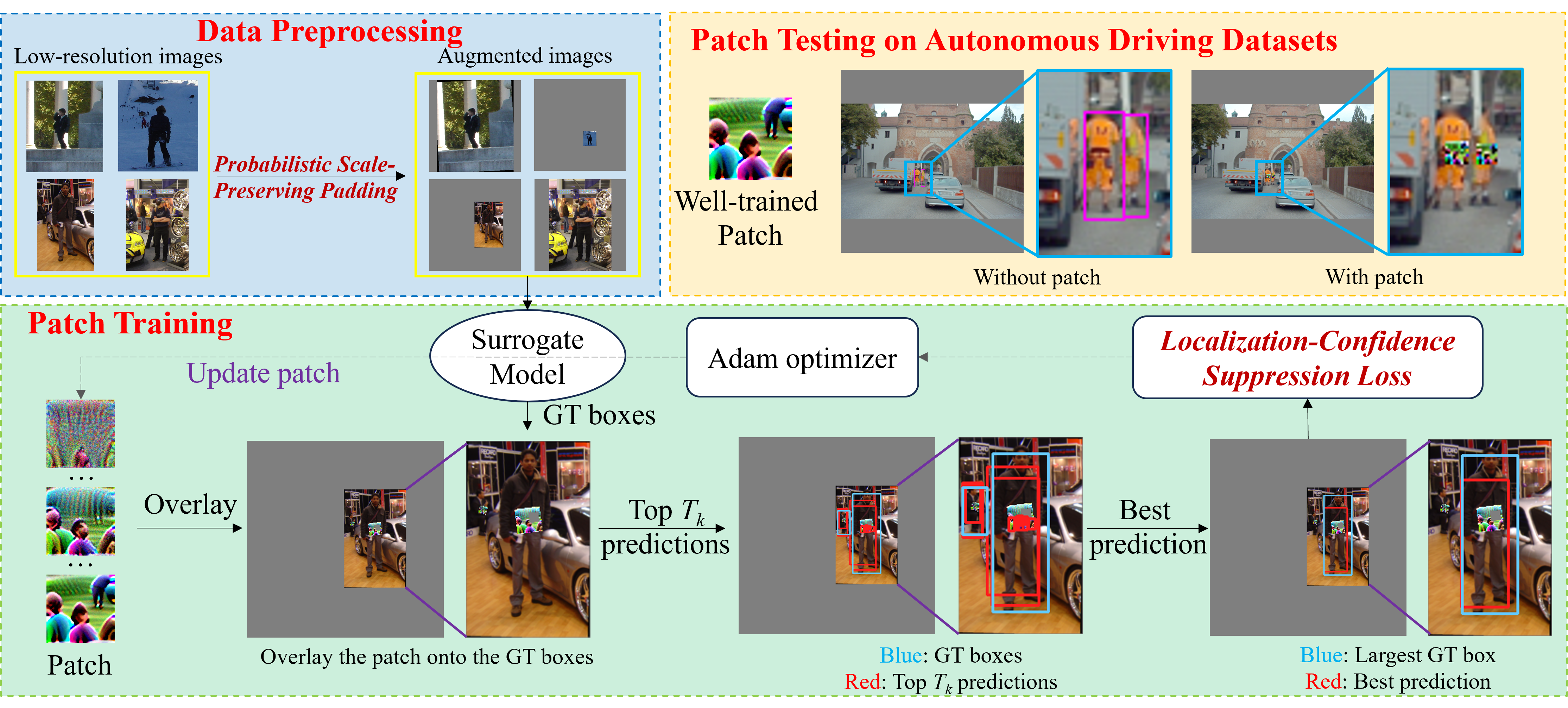}
    \caption{Pipeline of \ours, which primarily includes a probabilistic data preprocessing for HR data and a patch training stage using our designed adversarial loss.}
    \label{fig:framework}
\end{figure*}

\noindent\textbf{LCSL design.} Based on these analyses, we intuitively deem that incorporating IoU into score-based adversarial losses enables simultaneous suppression of both IoU and confidence score, thereby improving the effectiveness of the trained patch in terms of PASR and enhancing its generalization ability. 
Specifically, we first select the top $T_k$ predictions with the highest product of objectness and classification scores, \ie
\begin{equation}
\bar{\mathbf{O}}_{T_k} = \text{Top-}T_k \left( \left\{ \left( \bar{\mathbf{b}}_j, \bar{o}_j, \bar{\mathbf{s}}_j \right) \in \bar{\mathbf{O}} \,\middle|\, \bar{o}_j \cdot \bar{s}_j^0 \right\} \right),
\end{equation}
here we omit the adversarial image $\mathbf{x}_{adv}^i$ for simplicity, and $\bar{s}_j^0$ denotes the classification score for class 0 (person). For each of these boxes, we compute the IoU with $\mathbf{b}_{\max}^{\text{GT}}$, which refers to the one with the maximum area among all final GT boxes. This is motivated by the fact that, in autonomous driving scenarios, a larger person box typically indicates a closer distance to the vehicle, and thus missing such a detection poses a higher safety risk. Moreover, since our objective is to make at least one person disappear from the detector's output, attacking the largest GT box is both sufficient and computationally efficient. We then multiply the score product by the corresponding IoU, and the maximum of these $T_k$ final products is used to build our proposed LCSL, \ie
\begin{equation}\label{equ:adv_loss}
\mathcal{L}_{\text{adv}} = \frac{1}{N_b} \sum\nolimits_{i=1}^{N_b} \max_{(\bar{\mathbf{b}}_j, \bar{o}_j, \bar{\mathbf{s}}_j) \in \bar{\mathbf{O}}_{T_k}} \left[ \bar{o}_j \cdot \bar{s}_j^0 \cdot \text{IoU}\left( \bar{\mathbf{b}}_j, \mathbf{b}_{\max}^{\text{GT}} \right) \right].
\end{equation}
where $N_b$ denotes the image number in a mini-batch. To support the design of LCSL, we also provide a theoretical analysis based on empirical Rademacher complexity~\cite{yin2019rademacher}, as detailed in Appendix.

\subsection{Pipeline}
The overall pipeline of our proposed \ours is illustrated in Figure~\ref{fig:framework}, and the corresponding pseudo code is provided in Algorithm~\ref{alg:ours}. The total loss is formulated as a weighted sum of the adversarial loss and the total variation loss, where the latter, scaled by a loss weight $\lambda_{tv}$, encourages smoothness in the generated patch.

\setlength{\algomargin}{4pt}
\begin{algorithm}[h]\small
\footnotesize
\caption{Pseudo code of \ours.}\label{alg:ours}
\KwIn{
White-box surrogate detector $D_w$,
training images $X$, 
size of mini-batch $N_b$,
initial patch $\xi_0$, patch applier $T$, total variance loss $\mathcal{L}_\text{tv}$, maximum epoch $M$, learning rate $\eta$, preprocessing probability $p_{hr}$, preprocessing resolution $\left( H_{hr}, W_{hr} \right)$,
top number $T_k$, loss weight $\lambda_{tv}$.
}
\KwOut{Optimized patch $\xi$.}
$\xi \leftarrow \xi_0$;\\
\For{$i = 1$ \textbf{to} $M$}{
    \For{a mini-batch ${X_b}$ \textbf{in} $X$}{
        $S \leftarrow \emptyset$;\\
        \For{each $\mathbf{x}$ in $\mathbf{X_b}$}{
            $r \sim \text{Uniform}(0,1)$;\\
            \If{$r < p_{hr}$}{
                $x \leftarrow \text{Pad}\left(\mathbf{x}, H_{hr}, W_{hr}\right)$;//High resolution padding.\\
            }
            $\mathbf{b},{p},{c} \leftarrow \text{NMS} \left( D_w \left( \mathbf{x} \right) \right)$;\\
            $\mathbf{b}_{\max}^{\text{GT}} \leftarrow  \max \left( \mathbf{b}  \right)$;\\
            $\mathbf{x}_{adv} \leftarrow T \left( \mathbf{x}, \mathbf{b}, \xi \right)$;\\
            $\bar{\mathbf{O}} \leftarrow D_w(\mathbf{x}_{adv})$;\\
            $\bar{\mathbf{O}}_{T_k} \leftarrow \text{Top-}T_k \left( \left\{ \left( \bar{\mathbf{b}}_j, \bar{o}_j, \bar{\mathbf{s}}_j \right) \in \bar{\mathbf{O}} \,\middle|\, \bar{o}_j \cdot \bar{s}_j^0 \right\} \right)$;\\
            $S \leftarrow S \cup \left\{ \max\limits_{(\bar{\mathbf{b}}_j, \bar{o}_j, \bar{\mathbf{s}}_j) \in \bar{\mathbf{O}}_{T_k}} \left( \bar{o}_j \cdot \bar{s}_j^0 \cdot \text{IoU}\left( \bar{\mathbf{b}}_j, \mathbf{b}_{\max}^{\text{GT}} \right) \right) \right\}$;\\
        }
        $\mathcal{L}_{adv} \leftarrow \frac{1}{N_b} \sum_{m \in S} m$;\\
        $\mathcal{L}_{total} \leftarrow \mathcal{L}_{adv} + \lambda_{tv} \mathcal{L}_\text{tv}$;\\
        $\xi \leftarrow \xi - \eta {\nabla _\xi} \left( {\mathcal{L}_{total} } \right)$;\\
        }
    }
\end{algorithm}

\noindent\textbf{Data Preprocessing.}
To improve the transferability of adversarial patches to HR autonomous driving datasets, we introduce a HR preprocessing step, called Probabilistic Scale-Preserving Padding (PSPP). In autonomous driving datasets, the relative area of pedestrians with respect to the full image is much smaller, making it more challenging for the patch to remain effective. 
Since the attacker cannot access HR data during training, we simulate this scenario by padding LR training images to a predefined high resolution $H_{hr} \times W_{hr}$ with a probability $p_{hr}$. This strategy preserves the absolute size of pedestrians while reducing their relative size in the image, which better simulates the characteristics of autonomous driving scenes. It is worth noting that squarely resizing images to a high resolution would enlarge pedestrian sizes and fail to reflect real-world conditions realistically.

\noindent\textbf{Patch Training.}
As mentioned earlier, \tsea achieves a self-ensemble attack by jointly applying data augmentation, model ShakeDrop, and patch cutout during training. Following this design, we adopt the same strategy to enable our patch to achieve strong transferability under minimal training cost (only one surrogate required). The main difference is that we replace their score-based adversarial loss with our designed ones (Equation~\ref{equ:adv_loss}). Our loss selects the bounding box with high confidence scores and large IoU and minimizes their product to enhance transferability. 
Finally, the transferability of the trained patches is evaluated in terms of PASR across various detectors and multiple datasets.

\section{Experiments}
\subsection{Experimental Setup}
\noindent\textbf{Object Detectors.} We select 11 object detectors. For fair comparison, we use eight widely adopted models: YOLO v2~\cite{redmon2017yolo9000}, YOLO v3, YOLO v3tiny~\cite{redmon2018yolov3}, YOLO v4, YOLO v4tiny ~\cite{bochkovskiy2020yolov4}, YOLO v5~\cite{glenn_jocher_2020_4154370}, Faster R-CNN~\cite{ren2015faster} and SSD~\cite{liu2016ssd}. We further test on three recent detectors: RT-DETR2~\cite{lv2024Rt-detrv2}, YOLOv8~\cite{yolov8_2023ultralytics} and YOLO11~\cite{yolo11_2024ultralytics}.
These detectors cover both one-stage (\eg YOLO series and SSD) and two-stage (\eg Faster R-CNN) architectures, and represent a diverse set of models commonly used in real-world object detection tasks.

\noindent\textbf{Datasets.} We totally involve nine datasets in our experiments, comprising two LR person datasets from daily scenarios: 
INRIA-person~\cite{dalal2005inria} and COCO-person~\cite{lin2014coco} (hereafter INRIA and COCO), 
as well as seven HR datasets from autonomous driving scenarios: KITTI-i \& KITTI-p~\cite{geiger2012kitti}, bdd10k \& bdd100k~\cite{yu2020bdd100k}, nuScenes \& nuImages~\cite{caesar2020nuscenes} and A2D2~\cite{geyer2020a2d2}. 
KITTI-i refers to the standard KITTI dataset, while KITTI-p contains 3 temporally preceding frames. bdd10k is a subset of bdd100k. Both nuScenes and nuImages were collected by Motional, an autonomous driving company. nuImages is derived from nuScenes, focusing only on camera-based tasks. A2D2 was collected by Audi, featuring diverse urban and suburban scenes with pixel-level semantic annotations and bounding boxes for object detection tasks. We only use the camera data from all datasets. 

Table~\ref{tab:datasets} provides the basic information of these datasets. 
INRIA and COCO are specifically designed for person detection tasks, while the remaining seven datasets are autonomous driving datasets. Since not all images in these datasets contain persons, we filter the datasets using YOLO v3 by retaining only the images where at least one person is detected. We then calculate the number of selected images and their average resolution for each dataset. 
In terms of resolution, it is evident that the two person datasets have relatively low resolutions, approximately 600 pixels or lower, while the autonomous driving datasets have resolutions of at least 1,200 pixels, with the highest reaching 1,920 pixels. In these images, pedestrians are often small, and only when vehicles are relatively close to pedestrians does the pedestrian size become larger. This better reflects the real-world scenarios captured by autonomous vehicle sensors, where pedestrians are typically small in the field of view. However, we observe that current patch attacks on person detection primarily focus on the first two datasets or other datasets with similar resolution and data distribution, overlooking their limitations in transferring to HR datasets that more closely resemble real-world physical scenarios. Since the effectiveness of patch attacks tends to degrade when pedestrian size is small, we aim to solve this problem and enhance transferability when the patch is trained on LR data.

\begin{table}[!t]  
\centering
\caption{Dataset introduction. ``Num. of Images'' refers to the count of images where YOLO v3 detects at least one person.
}
\label{tab:datasets}
\begin{tabular}{ccc}
\toprule
Dataset & Num. of Images & Average Resolution
 \\
\midrule
INRIA & 288 & (630.7, 647.0) \\
COCO & 475 & (573.9, 485.0) \\
KITTI-i & 1,769 & (1,233.6, 372.6) \\
KITTI-p & 5,258 & (1,233.5, 372.6) \\
bdd10k & 194 & (1,280.0, 720.0) \\
bdd100k & 1,689 & (1,280.0, 720.0) \\
nuScene & 7,804 & (1,600.0, 900.0) \\
nuImages & 5,134 & (1,600.0, 900.0) \\
A2D2 & 1,968 & (1,920.0, 1,208.0) \\
\bottomrule
\end{tabular}
\end{table}

\noindent\textbf{Attack Metrics.}
We use the proposed PASR as the main metric. We also provide results of the two traditional metrics, mAP and ASR, as a reference.

\noindent\textbf{Competitors.}
Since our method is structurally based on \tsea, we treat \tsea as our baseline and conduct a comprehensive comparison with it in both cross-model and cross-dataset experiments. Additionally, we compare our method with 13 existing patch attack methods: AdvPatch~\cite{thys2019advpatch}, AdvTshirt~\cite{xu2020advtshirt}, AdvCloak~\cite{wu2020advcloak}, AdvTexture~\cite{hu2022advtexture}, NAP~\cite{hu2021naturalistic}, LAP~\cite{tan2021lap}, DAP~\cite{guesmi2024dap}, CAP~\cite{wei2024cap}, FDA~\cite{cheng2024fda}, DePatch~\cite{cheng2024depatch}, MVPatch~\cite{zhou2023mvpatch}, AdvART~\cite{guesmi2024advart}, and \tsea~\cite{huang2023tsea}.

\begin{table*}[t]  
\centering
\caption{Attack performance of cross-model transferability. Imp. denotes improvement. \textcolor{blue}{Blue} indicates better performance of our attack; \textcolor{red}{red} denotes worse.}
\label{tab:attack_performance_cross_model}
\resizebox{0.97\linewidth}{!}{
\begin{tabular}{ccrrrrrrrrrrrrrrrrrrrrrrrrrrr}
\toprule
\multirow{2}{*}[-0.5ex]{\diagbox{Train}{Test}} & 
\multirow{2}{*}[-0.5ex]{Method} & \multicolumn{3}{c}{YOLO v2} & \multicolumn{3}{c}{YOLO v3}  & \multicolumn{3}{c}{YOLO v3tiny} & \multicolumn{3}{c}{YOLO v4}  & \multicolumn{3}{c}{YOLO v4tiny} & \multicolumn{3}{c}{YOLO v5}  & \multicolumn{3}{c}{Faster-RCNN} & \multicolumn{3}{c}{SSD} \\
\cmidrule(r){3-5}\cmidrule(r){6-8}\cmidrule(r){9-11}\cmidrule(r){12-14}\cmidrule(r){15-17}\cmidrule(r){18-20}\cmidrule(r){21-23}\cmidrule(r){24-26}
& & \footnotesize{PASR$\uparrow$} & \footnotesize{mAP$\downarrow$} & \footnotesize{ASR$\uparrow$} & \footnotesize{PASR$\uparrow$} & \footnotesize{mAP$\downarrow$} & \footnotesize{ASR$\uparrow$} & \footnotesize{PASR$\uparrow$} & \footnotesize{mAP$\downarrow$} & \footnotesize{ASR$\uparrow$} & \footnotesize{PASR$\uparrow$} & \footnotesize{mAP$\downarrow$} & \footnotesize{ASR$\uparrow$} & \footnotesize{PASR$\uparrow$} & \footnotesize{mAP$\downarrow$} & \footnotesize{ASR$\uparrow$} & \footnotesize{PASR$\uparrow$} & \footnotesize{mAP$\downarrow$} & \footnotesize{ASR$\uparrow$} & \footnotesize{PASR$\uparrow$} & \footnotesize{mAP$\downarrow$} & \footnotesize{ASR$\uparrow$} & \footnotesize{PASR$\uparrow$} & \footnotesize{mAP$\downarrow$} & \footnotesize{ASR$\uparrow$} \\
\midrule
\multirow{3}{*}{YOLO v2} & \tsea  
& 67.18 & 20.71 & 73.26 & 36.16 & 32.22 & 52.28 & 50.95 & 31.71 & 60.52 & 23.16 & 67.17 & 26.41 & 44.11 & 44.98 & 48.95 & 31.67 & 63.96 & 31.54 & 26.03 & 53.39 & 32.41 & 35.18 & 57.28 & 38.86 \\
& \ours 
& 71.28 & 12.83 & 79.95 & 50.24 & 16.45 & 67.40 & 55.53 & 25.84 & 66.22 & 31.20 & 55.13 & 34.18 & 50.52 & 30.79 & 59.64 & 35.74 & 56.35 & 36.42 & 47.46 & 27.00 & 52.55 & 49.50 & 35.64 & 57.60 \\
& Imp.
& \textcolor{blue}{4.10} & \textcolor{blue}{7.88} & \textcolor{blue}{6.69} & \textcolor{blue}{14.08} & \textcolor{blue}{15.77} & \textcolor{blue}{15.12} & \textcolor{blue}{4.58} & \textcolor{blue}{5.87} & \textcolor{blue}{5.70} & \textcolor{blue}{8.04} & \textcolor{blue}{12.04} & \textcolor{blue}{7.77} & \textcolor{blue}{6.41} & \textcolor{blue}{14.19} & \textcolor{blue}{10.69} & \textcolor{blue}{4.07} & \textcolor{blue}{7.61} & \textcolor{blue}{4.88} & \textcolor{blue}{21.43} & \textcolor{blue}{26.39} & \textcolor{blue}{20.14} & \textcolor{blue}{14.32} & \textcolor{blue}{21.64} & \textcolor{blue}{18.74} \\

\midrule
\multirow{3}{*}{YOLO v3} & \tsea 
& 54.34 & 40.52 & 55.53 & 58.36 & 35.07 & 59.51 & 63.15 & 35.81 & 62.50 & 35.34 & 68.69 & 30.32 & 53.50 & 47.62 & 51.29 & 45.65 & 59.81 & 39.58 & 47.77 & 53.40 & 42.35 & 43.85 & 56.57 & 42.60 \\
& \ours 
& 60.12 & 29.72 & 64.54 & 60.38 & 22.30 & 64.17 & 69.02 & 22.90 & 72.37 & 36.82 & 58.70 & 35.00 & 59.14 & 34.10 & 61.64 & 49.00 & 51.26 & 44.74 & 48.92 & 43.36 & 46.48 & 47.58 & 50.42 & 47.51 \\
& Imp.
& \textcolor{blue}{5.78} & \textcolor{blue}{10.80} & \textcolor{blue}{9.01} & \textcolor{blue}{2.02} & \textcolor{blue}{12.77} & \textcolor{blue}{4.66} & \textcolor{blue}{5.87} & \textcolor{blue}{12.91} & \textcolor{blue}{9.87} & \textcolor{blue}{1.48} & \textcolor{blue}{9.99} & \textcolor{blue}{4.68} & \textcolor{blue}{5.64} & \textcolor{blue}{13.52} & \textcolor{blue}{10.35} & \textcolor{blue}{3.35} & \textcolor{blue}{8.55} & \textcolor{blue}{5.16} & \textcolor{blue}{1.15} & \textcolor{blue}{10.04} & \textcolor{blue}{4.13} & \textcolor{blue}{3.73} & \textcolor{blue}{6.15} & \textcolor{blue}{4.91} \\

\midrule
\multirow{3}{*}{YOLO v3tiny} & \tsea 
& 59.96 & 33.03 & 61.47 & 52.17 & 36.79 & 56.57 & 80.09 & 10.83 & 85.28 & 28.51 & 72.44 & 25.20 & 51.57 & 44.49 & 51.09 & 41.08 & 63.40 & 35.51 & 46.36 & 52.45 & 41.90 & 41.70 & 55.88 & 41.58 \\
& \ours 
& 67.76 & 24.37 & 71.40 & 59.15 & 29.79 & 64.25 & 75.34 & 3.21 & 92.09 & 33.35 & 66.18 & 30.65 & 63.58 & 29.00 & 66.73 & 44.43 & 58.44 & 39.57 & 53.92 & 42.94 & 50.53 & 46.25 & 49.75 & 47.52 \\
& Imp.
& \textcolor{blue}{7.80} & \textcolor{blue}{8.66} & \textcolor{blue}{9.93} & \textcolor{blue}{6.98} & \textcolor{blue}{7.00} & \textcolor{blue}{7.68} & \textcolor{red}{-4.75} & \textcolor{blue}{7.62} & \textcolor{blue}{6.81} & \textcolor{blue}{4.84} & \textcolor{blue}{6.26} & \textcolor{blue}{5.45} & \textcolor{blue}{12.01} & \textcolor{blue}{15.49} & \textcolor{blue}{15.64} & \textcolor{blue}{3.35} & \textcolor{blue}{4.96} & \textcolor{blue}{4.06} & \textcolor{blue}{7.56} & \textcolor{blue}{9.51} & \textcolor{blue}{8.63} & \textcolor{blue}{4.55} & \textcolor{blue}{6.13} & \textcolor{blue}{5.94} \\

\midrule
\multirow{3}{*}{YOLO v4} & \tsea 
& 56.54 & 29.02 & 62.24 & 35.97 & 28.32 & 52.45 & 49.06 & 35.70 & 53.32 & 23.78 & 65.37 & 27.67 & 40.35 & 47.79 & 43.50 & 32.21 & 62.98 & 32.21 & 25.38 & 55.06 & 30.99 & 32.72 & 62.00 & 35.01 \\
& \ours 
& 58.47 & 28.04 & 63.83 & 41.29 & 30.56 & 54.36 & 54.26 & 32.39 & 61.65 & 25.68 & 62.58 & 31.73 & 39.47 & 47.76 & 44.77 & 35.60 & 60.36 & 34.48 & 31.42 & 48.66 & 34.57 & 37.85 & 54.13 & 40.75 \\
& Imp.
& \textcolor{blue}{1.93} & \textcolor{blue}{0.98} & \textcolor{blue}{1.59} & \textcolor{blue}{5.32} & \textcolor{red}{-2.24} & \textcolor{blue}{1.91} & \textcolor{blue}{5.20} & \textcolor{blue}{3.31} & \textcolor{blue}{8.33} & \textcolor{blue}{1.90} & \textcolor{blue}{2.79} & \textcolor{blue}{4.06} & \textcolor{red}{-0.88} & \textcolor{blue}{0.03} & \textcolor{blue}{1.27} & \textcolor{blue}{3.39} & \textcolor{blue}{2.62} & \textcolor{blue}{2.27} & \textcolor{blue}{6.04} & \textcolor{blue}{6.40} & \textcolor{blue}{3.58} & \textcolor{blue}{5.13} & \textcolor{blue}{7.87} & \textcolor{blue}{5.74} \\

\midrule
\multirow{3}{*}{YOLO v4tiny} & \tsea 
& 57.50 & 36.05 & 60.14 & 47.89 & 35.14 & 55.97 & 57.63 & 35.91 & 60.33 & 24.66 & 74.56 & 22.86 & 51.29 & 43.16 & 53.53 & 35.04 & 67.91 & 30.77 & 45.69 & 51.84 & 41.77 & 39.61 & 59.03 & 38.93 \\
& \ours 
& 61.55 & 33.29 & 62.14 & 56.73 & 40.64 & 55.99 & 72.21 & 29.03 & 70.04 & 36.01 & 69.19 & 30.21 & 68.98 & 32.39 & 66.92 & 47.04 & 59.11 & 40.28 & 59.04 & 42.65 & 54.09 & 60.18 & 39.20 & 59.66 \\
& Imp.
& \textcolor{blue}{4.05} & \textcolor{blue}{2.76} & \textcolor{blue}{2.00} & \textcolor{blue}{8.84} & \textcolor{red}{-5.50} & \textcolor{blue}{0.02} & \textcolor{blue}{14.58} & \textcolor{blue}{6.88} & \textcolor{blue}{9.71} & \textcolor{blue}{11.35} & \textcolor{blue}{5.37} & \textcolor{blue}{7.35} & \textcolor{blue}{17.69} & \textcolor{blue}{10.77} & \textcolor{blue}{13.39} & \textcolor{blue}{12.00} & \textcolor{blue}{8.80} & \textcolor{blue}{9.51} & \textcolor{blue}{13.35} & \textcolor{blue}{9.19} & \textcolor{blue}{12.32} & \textcolor{blue}{20.57} & \textcolor{blue}{19.83} & \textcolor{blue}{20.73} \\

\midrule
\multirow{3}{*}{YOLO v5} & \tsea 
& 50.46 & 43.96 & 52.69 & 50.98 & 44.68 & 52.22 & 58.98 & 42.28 & 56.30 & 25.28 & 78.42 & 21.34 & 41.44 & 62.11 & 37.68 & 40.77 & 64.10 & 35.88 & 41.48 & 62.50 & 35.24 & 40.32 & 60.44 & 40.14 \\
& \ours 
& 50.76 & 38.64 & 55.68 & 48.46 & 44.40 & 49.37 & 61.05 & 33.18 & 63.26 & 35.07 & 64.08 & 32.17 & 53.92 & 44.18 & 53.12 & 45.29 & 54.40 & 41.58 & 48.35 & 47.62 & 45.92 & 46.62 & 52.07 & 46.34 \\
& Imp.
& \textcolor{blue}{0.30} & \textcolor{blue}{5.32} & \textcolor{blue}{2.99} & \textcolor{red}{-2.52} & \textcolor{blue}{0.28} & \textcolor{red}{-2.85} & \textcolor{blue}{2.07} & \textcolor{blue}{9.10} & \textcolor{blue}{6.96} & \textcolor{blue}{9.79} & \textcolor{blue}{14.34} & \textcolor{blue}{10.83} & \textcolor{blue}{12.48} & \textcolor{blue}{17.93} & \textcolor{blue}{15.44} & \textcolor{blue}{4.52} & \textcolor{blue}{9.70} & \textcolor{blue}{5.70} & \textcolor{blue}{6.87} & \textcolor{blue}{14.88} & \textcolor{blue}{10.68} & \textcolor{blue}{6.30} & \textcolor{blue}{8.37} & \textcolor{blue}{6.20} \\

\midrule
\multirow{3}{*}{Faster-RCNN} & \tsea 
& 41.24 & 53.19 & 43.21 & 42.28 & 54.19 & 41.22 & 56.88 & 43.34 & 55.16 & 33.44 & 69.94 & 28.27 & 46.65 & 55.40 & 43.75 & 47.15 & 57.74 & 41.05 & 42.69 & 53.86 & 38.69 & 39.07 & 61.54 & 37.32 \\
& \ours 
& 53.68 & 35.03 & 58.92 & 54.60 & 30.25 & 61.11 & 73.72 & 23.08 & 74.29 & 34.47 & 64.18 & 31.76 & 55.63 & 43.62 & 54.50 & 47.95 & 54.19 & 43.35 & 55.99 & 34.69 & 55.32 & 54.17 & 41.82 & 55.66 \\
& Imp.
& \textcolor{blue}{12.44} & \textcolor{blue}{18.16} & \textcolor{blue}{15.71} & \textcolor{blue}{12.32} & \textcolor{blue}{23.94} & \textcolor{blue}{19.89} & \textcolor{blue}{16.84} & \textcolor{blue}{20.26} & \textcolor{blue}{19.13} & \textcolor{blue}{1.03} & \textcolor{blue}{5.76} & \textcolor{blue}{3.49} & \textcolor{blue}{8.98} & \textcolor{blue}{11.78} & \textcolor{blue}{10.75} & \textcolor{blue}{0.80} & \textcolor{blue}{3.55} & \textcolor{blue}{2.30} & \textcolor{blue}{13.30} & \textcolor{blue}{19.17} & \textcolor{blue}{16.63} & \textcolor{blue}{15.10} & \textcolor{blue}{19.72} & \textcolor{blue}{18.34} \\

\midrule
\multirow{3}{*}{SSD} & \tsea 
& 61.89 & 34.33 & 62.61 & 55.66 & 37.36 & 58.19 & 71.92 & 26.44 & 70.60 & 35.72 & 68.14 & 30.94 & 60.87 & 38.20 & 58.99 & 45.05 & 60.12 & 38.80 & 47.32 & 48.65 & 46.89 & 47.41 & 52.31 & 46.70 \\
& \ours 
& 63.00 & 34.81 & 62.67 & 69.93 & 29.76 & 68.10 & 85.67 & 14.78 & 84.06 & 43.55 & 63.39 & 36.22 & 66.66 & 37.05 & 62.77 & 48.67 & 58.90 & 40.92 & 62.83 & 41.72 & 56.43 & 51.04 & 51.09 & 48.65 \\
& Imp.
& \textcolor{blue}{1.11} & \textcolor{red}{-0.48} & \textcolor{blue}{0.06} & \textcolor{blue}{14.27} & \textcolor{blue}{7.60} & \textcolor{blue}{9.91} & \textcolor{blue}{13.75} & \textcolor{blue}{11.66} & \textcolor{blue}{13.46} & \textcolor{blue}{7.83} & \textcolor{blue}{4.75} & \textcolor{blue}{5.28} & \textcolor{blue}{5.79} & \textcolor{blue}{1.15} & \textcolor{blue}{3.78} & \textcolor{blue}{3.62} & \textcolor{blue}{1.22} & \textcolor{blue}{2.12} & \textcolor{blue}{15.51} & \textcolor{blue}{6.93} & \textcolor{blue}{9.54} & \textcolor{blue}{3.63} & \textcolor{blue}{1.22} & \textcolor{blue}{1.95} \\

\bottomrule
\end{tabular}
}
\end{table*}

\begin{table*}[t]  
\centering
\caption{Attack performance of cross-dataset transferability (HR data). Imp. denotes improvement. \textcolor{blue}{Blue} indicates better performance of our attack; \textcolor{red}{red} denotes worse.}
\label{tab:attack_performance_hr_transfer}
\resizebox{0.98\linewidth}{!}{
\begin{tabular}{ccrrrrrrrrrrrrrrrrrrrrrrrrrrrrrr}
\toprule
\multirow{2}{*}[-0.5ex]{\diagbox{Train}{Test}} & 
\multirow{2}{*}[-0.5ex]{Method} & \multicolumn{3}{c}{KITTI-i} & \multicolumn{3}{c}{KITTI-p}  & \multicolumn{3}{c}{bdd10k} & \multicolumn{3}{c}{bdd100k}  & \multicolumn{3}{c}{nuScenes} & \multicolumn{3}{c}{nuImages} & \multicolumn{3}{c}{A2D2} & \multicolumn{3}{c}{\textbf{Avg}} \\
\cmidrule(r){3-5}\cmidrule(r){6-8}\cmidrule(r){9-11}\cmidrule(r){12-14}\cmidrule(r){15-17}\cmidrule(r){18-20}\cmidrule(r){21-23}\cmidrule(r){24-26}\cmidrule(r){27-29}\cmidrule(r){30-32}
& & \footnotesize{PASR$\uparrow$} & \footnotesize{mAP$\downarrow$} & \footnotesize{ASR$\uparrow$} & \footnotesize{PASR$\uparrow$} & \footnotesize{mAP$\downarrow$} & \footnotesize{ASR$\uparrow$} & \footnotesize{PASR$\uparrow$} & \footnotesize{mAP$\downarrow$} & \footnotesize{ASR$\uparrow$} & \footnotesize{PASR$\uparrow$} & \footnotesize{mAP$\downarrow$} & \footnotesize{ASR$\uparrow$} & \footnotesize{PASR$\uparrow$} & \footnotesize{mAP$\downarrow$} & \footnotesize{ASR$\uparrow$} & \footnotesize{PASR$\uparrow$} & \footnotesize{mAP$\downarrow$} & \footnotesize{ASR$\uparrow$} & \footnotesize{PASR$\uparrow$} & \footnotesize{mAP$\downarrow$} & \footnotesize{ASR$\uparrow$} & \footnotesize{PASR$\uparrow$} & \footnotesize{mAP$\downarrow$} & \footnotesize{ASR$\uparrow$} \\
\midrule
\multirow{3}{*}{YOLO v2} &
\tsea
& 28.58 & 63.97 & 31.11 & 28.31 & 64.13 & 30.81 & 32.80 & 49.08 & 41.53 & 46.96 & 40.43 & 49.63 & 39.59 & 44.45 & 46.09 & 40.94 & 47.24 & 45.18 & 34.41 & 55.73 & 35.95 & 35.94 & 52.15 & 40.04 \\
& \ours
& 37.57 & 51.15 & 41.13 & 37.63 & 50.69 & 41.36 & 45.75 & 33.12 & 55.23 & 58.78 & 26.00 & 62.61 & 53.42 & 29.16 & 60.20 & 55.83 & 30.87 & 60.27 & 44.96 & 41.08 & 47.76 & 47.71 & 37.44 & 52.65 \\
& Imp.
& \textcolor{blue}{8.99} & \textcolor{blue}{12.82} & \textcolor{blue}{10.02} & \textcolor{blue}{9.32} & \textcolor{blue}{13.44} & \textcolor{blue}{10.55} & \textcolor{blue}{12.95} & \textcolor{blue}{15.96} & \textcolor{blue}{13.70} & \textcolor{blue}{11.82} & \textcolor{blue}{14.43} & \textcolor{blue}{12.98} & \textcolor{blue}{13.83} & \textcolor{blue}{15.29} & \textcolor{blue}{14.11} & \textcolor{blue}{14.89} & \textcolor{blue}{16.37} & \textcolor{blue}{15.09} & \textcolor{blue}{10.55} & \textcolor{blue}{14.65} & \textcolor{blue}{11.81} & \textcolor{blue}{11.77} & \textcolor{blue}{14.71} & \textcolor{blue}{12.61} \\

\midrule
\multirow{3}{*}{YOLO v3} &
\tsea
& 37.11 & 65.64 & 33.57 & 37.79 & 64.74 & 34.31 & 50.96 & 48.31 & 49.72 & 64.39 & 40.52 & 57.51 & 55.45 & 47.04 & 51.30 & 58.38 & 45.21 & 53.72 & 49.74 & 54.95 & 43.34 & 50.55 & 52.34 & 46.21 \\
& \ours
& 45.04 & 57.66 & 41.32 & 45.09 & 57.62 & 41.23 & 56.45 & 38.51 & 56.20 & 67.66 & 31.11 & 63.84 & 60.65 & 37.90 & 58.21 & 63.18 & 37.84 & 59.88 & 53.85 & 47.04 & 49.42 & 55.99 & 43.95 & 52.87 \\
& Imp.
& \textcolor{blue}{7.93} & \textcolor{blue}{7.98} & \textcolor{blue}{7.75} & \textcolor{blue}{7.30} & \textcolor{blue}{7.12} & \textcolor{blue}{6.92} & \textcolor{blue}{5.49} & \textcolor{blue}{9.80} & \textcolor{blue}{6.48} & \textcolor{blue}{3.27} & \textcolor{blue}{9.41} & \textcolor{blue}{6.33} & \textcolor{blue}{5.20} & \textcolor{blue}{9.14} & \textcolor{blue}{6.91} & \textcolor{blue}{4.80} & \textcolor{blue}{7.37} & \textcolor{blue}{6.16} & \textcolor{blue}{4.11} & \textcolor{blue}{7.91} & \textcolor{blue}{6.08} & \textcolor{blue}{5.44} & \textcolor{blue}{8.39} & \textcolor{blue}{6.66} \\

\midrule
\multirow{3}{*}{YOLO v3tiny} &
\tsea
& 38.57 & 62.53 & 35.93 & 38.75 & 62.08 & 36.22 & 48.66 & 49.39 & 47.68 & 60.58 & 40.91 & 55.34 & 52.52 & 46.56 & 50.10 & 55.38 & 44.45 & 52.84 & 46.31 & 55.97 & 40.93 & 48.68 & 51.70 & 45.58 \\
& \ours
& 43.86 & 57.03 & 41.25 & 44.11 & 56.70 & 41.51 & 56.53 & 38.80 & 57.01 & 67.30 & 30.39 & 64.60 & 61.92 & 35.83 & 60.14 & 64.57 & 34.45 & 62.52 & 52.94 & 48.22 & 48.21 & 55.89 & 43.06 & 53.61 \\
& Imp.
& \textcolor{blue}{5.29} & \textcolor{blue}{5.50} & \textcolor{blue}{5.32} & \textcolor{blue}{5.36} & \textcolor{blue}{5.38} & \textcolor{blue}{5.29} & \textcolor{blue}{7.87} & \textcolor{blue}{-10.59} & \textcolor{blue}{9.33} & \textcolor{blue}{6.72} & \textcolor{blue}{-10.52} & \textcolor{blue}{9.26} & \textcolor{blue}{9.40} & \textcolor{blue}{-10.73} & \textcolor{blue}{10.04} & \textcolor{blue}{9.19} & \textcolor{blue}{10.00} & \textcolor{blue}{9.68} & \textcolor{blue}{6.63} & \textcolor{blue}{7.75} & \textcolor{blue}{7.28} & \textcolor{blue}{7.21} & \textcolor{blue}{-8.64} & \textcolor{blue}{8.03} \\

\midrule
\multirow{3}{*}{YOLO v4} &
\tsea
& 33.12 & 59.67 & 35.59 & 33.78 & 58.79 & 36.13 & 39.23 & 43.96 & 46.89 & 49.94 & 36.71 & 53.81 & 43.05 & 42.04 & 48.90 & 45.04 & 43.38 & 49.46 & 36.47 & 51.53 & 40.02 & 40.09 & 48.01 & 44.40 \\
& \ours
& 36.94 & 56.99 & 38.30 & 35.73 & 57.65 & 37.52 & 44.62 & 42.68 & 49.24 & 55.00 & 34.40 & 56.21 & 49.14 & 40.13 & 50.51 & 52.53 & 37.48 & 54.99 & 40.17 & 51.44 & 41.43 & 44.88 & 45.82 & 46.89 \\
& Imp.
& \textcolor{blue}{3.82} & \textcolor{blue}{2.68} & \textcolor{blue}{2.71} & \textcolor{blue}{1.95} & \textcolor{blue}{1.14} & \textcolor{blue}{1.39} & \textcolor{blue}{5.39} & \textcolor{blue}{1.28} & \textcolor{blue}{2.35} & \textcolor{blue}{5.06} & \textcolor{blue}{2.31} & \textcolor{blue}{2.40} & \textcolor{blue}{6.09} & \textcolor{blue}{1.91} & \textcolor{blue}{1.61} & \textcolor{blue}{7.49} & \textcolor{blue}{5.90} & \textcolor{blue}{5.53} & \textcolor{blue}{3.70} & \textcolor{blue}{0.09} & \textcolor{blue}{1.41} & \textcolor{blue}{4.79} & \textcolor{blue}{2.19} & \textcolor{blue}{2.49} \\

\midrule
\multirow{3}{*}{YOLO v4tiny} &
\tsea
& 34.32 & 65.42 & 32.67 & 34.87 & 64.23 & 33.66 & 44.87 & 50.67 & 45.64 & 58.75 & 40.69 & 55.02 & 51.37 & 45.13 & 50.72 & 53.59 & 44.00 & 52.74 & 43.21 & 56.27 & 39.99 & 45.85 & 52.34 & 44.35 \\
& \ours
& 44.30 & 59.21 & 39.72 & 42.56 & 59.17 & 38.52 & 62.21 & 39.31 & 58.24 & 70.88 & 33.04 & 65.18 & 65.71 & 36.55 & 61.72 & 68.09 & 34.40 & 64.18 & 58.35 & 46.17 & 51.71 & 58.87 & 43.98 & 54.18 \\
& Imp.
& \textcolor{blue}{9.98} & \textcolor{blue}{6.21} & \textcolor{blue}{7.05} & \textcolor{blue}{7.69} & \textcolor{blue}{5.06} & \textcolor{blue}{4.86} & \textcolor{blue}{17.34} & \textcolor{blue}{11.36} & \textcolor{blue}{12.60} & \textcolor{blue}{12.13} & \textcolor{blue}{7.65} & \textcolor{blue}{10.16} & \textcolor{blue}{14.34} & \textcolor{blue}{8.58} & \textcolor{blue}{11.00} & \textcolor{blue}{14.50} & \textcolor{blue}{9.60} & \textcolor{blue}{11.44} & \textcolor{blue}{15.14} & \textcolor{blue}{10.10} & \textcolor{blue}{11.72} & \textcolor{blue}{13.02} & \textcolor{blue}{8.36} & \textcolor{blue}{9.83} \\

\midrule
\multirow{3}{*}{YOLO v5} &
\tsea
& 29.48 & 72.47 & 27.09 & 29.89 & 71.78 & 27.53 & 45.09 & 54.15 & 44.83 & 56.37 & 47.95 & 50.58 & 54.13 & 47.87 & 51.56 & 56.66 & 45.98 & 53.60 & 44.70 & 59.01 & 39.86 & 45.19 & 57.03 & 42.15 \\
& \ours
& 37.11 & 64.21 & 34.58 & 37.05 & 64.09 & 34.51 & 52.98 & 41.85 & 53.64 & 63.63 & 33.94 & 61.28 & 58.83 & 38.12 & 57.69 & 61.86 & 37.38 & 59.89 & 50.86 & 49.49 & 46.46 & 51.76 & 47.01 & 49.72 \\
& Imp.
& \textcolor{blue}{7.63} & \textcolor{blue}{8.26} & \textcolor{blue}{7.49} & \textcolor{blue}{7.16} & \textcolor{blue}{7.69} & \textcolor{blue}{6.98} & \textcolor{blue}{7.89} & \textcolor{blue}{12.30} & \textcolor{blue}{8.81} & \textcolor{blue}{7.26} & \textcolor{blue}{14.01} & \textcolor{blue}{10.70} & \textcolor{blue}{4.70} & \textcolor{blue}{9.75} & \textcolor{blue}{6.13} & \textcolor{blue}{5.20} & \textcolor{blue}{8.60} & \textcolor{blue}{6.29} & \textcolor{blue}{6.16} & \textcolor{blue}{9.52} & \textcolor{blue}{6.60} & \textcolor{blue}{6.57} & \textcolor{blue}{10.02} & \textcolor{blue}{7.57} \\

\midrule
\multirow{3}{*}{Faster-RCNN} &
\tsea
& 34.12 & 68.46 & 30.86 & 34.95 & 67.80 & 31.30 & 46.61 & 52.67 & 44.98 & 57.26 & 44.89 & 52.38 & 49.96 & 51.74 & 46.20 & 54.12 & 48.56 & 49.86 & 44.22 & 59.64 & 38.25 & 45.89 & 56.25 & 41.98 \\
& \ours
& 43.70 & 56.90 & 41.64 & 44.24 & 56.17 & 42.18 & 55.53 & 39.49 & 56.46 & 67.69 & 30.64 & 65.31 & 64.54 & 33.73 & 63.07 & 68.34 & 31.39 & 66.55 & 51.44 & 48.71 & 47.93 & 56.50 & 42.43 & 54.73 \\
& Imp.
& \textcolor{blue}{9.58} & \textcolor{blue}{11.56} & \textcolor{blue}{10.78} & \textcolor{blue}{9.29} & \textcolor{blue}{11.63} & \textcolor{blue}{10.88} & \textcolor{blue}{8.92} & \textcolor{blue}{13.18} & \textcolor{blue}{11.48} & \textcolor{blue}{10.43} & \textcolor{blue}{14.25} & \textcolor{blue}{12.93} & \textcolor{blue}{14.58} & \textcolor{blue}{18.01} & \textcolor{blue}{16.87} & \textcolor{blue}{14.22} & \textcolor{blue}{17.17} & \textcolor{blue}{16.69} & \textcolor{blue}{7.22} & \textcolor{blue}{10.93} & \textcolor{blue}{9.68} & \textcolor{blue}{10.61} & \textcolor{blue}{13.82} & \textcolor{blue}{12.75} \\

\midrule
\multirow{3}{*}{SSD} &
\tsea
& 43.81 & 57.10 & 41.32 & 43.97 & 56.36 & 41.91 & 58.74 & 40.75 & 56.84 & 65.80 & 31.93 & 65.29 & 64.88 & 35.00 & 62.60 & 67.31 & 32.61 & 65.11 & 53.04 & 49.68 & 47.78 & 56.79 & 43.35 & 54.41 \\
& \ours
& 52.65 & 51.62 & 47.36 & 52.81 & 51.11 & 47.73 & 66.88 & 36.23 & 62.61 & 76.98 & 29.01 & 69.96 & 73.74 & 29.68 & 69.29 & 75.81 & 27.65 & 71.46 & 63.04 & 43.08 & 55.39 & 65.99 & 38.34 & 60.54 \\
& Imp.
& \textcolor{blue}{8.84} & \textcolor{blue}{5.48} & \textcolor{blue}{6.04} & \textcolor{blue}{8.84} & \textcolor{blue}{5.25} & \textcolor{blue}{5.82} & \textcolor{blue}{8.14} & \textcolor{blue}{4.52} & \textcolor{blue}{5.77} & \textcolor{blue}{11.18} & \textcolor{blue}{2.92} & \textcolor{blue}{4.67} & \textcolor{blue}{8.86} & \textcolor{blue}{5.32} & \textcolor{blue}{6.69} & \textcolor{blue}{8.50} & \textcolor{blue}{4.96} & \textcolor{blue}{6.35} & \textcolor{blue}{10.00} & \textcolor{blue}{6.60} & \textcolor{blue}{7.61} & \textcolor{blue}{9.20} & \textcolor{blue}{5.01} & \textcolor{blue}{6.13} \\

\bottomrule
\end{tabular}
}
\end{table*}

\noindent\textbf{Implementation.}
In our experiments, following previous work~\cite{huang2023tsea}, our patch is trained on a single detector using the training set of the INRIA dataset. 
Then we test the cross-model transferability and cross-dataset transferability on the other 10 detectors and the other eight datasets, respectively. Considering the resolution of our test datasets, we set the preprocessing resolution $\left( H_{hr}, W_{hr} \right)$ to (1,920, 1,920) as a reference value that achieves good transferability. Users may choose other resolutions according to their specific needs. We set the top number $T_k$ to 10. The preprocessing probability $p_{hr}$ is set to 0.5 based on a grid search. We adopt the Adam optimizer~\cite{kingma2014adam} with a scheduler and, for fair comparison, we keep the other parameters consistent with \tsea~\cite{huang2023tsea}, where the main settings include the maximum epoch $M=1,000$, learning rate $\eta=0.03$ and loss weight $\lambda_{tv}=2.5$. 

For competitors, we use the trained patches obtained from the tools of those papers when available; otherwise, we re-implement their methods based on the descriptions in the original papers. Then, we re-implement all the compared methods on our benchmark for evaluation. 

Following prior work such as \tsea and NAP, we select class 0 (person) as our attack target due to its high safety relevance in autonomous driving, where missing a pedestrian poses serious collision risks. But this application-driven choice does not mean our attack cannot be generalized to other object classes. Technically, extending either our method or existing attacks to other target classes (\eg vehicle, stop sign) is straightforward and involves only replacing the target label, and no change in architecture or training procedure is required.

\subsection{Attack Performance}

\subsubsection{Cross-Model Attacks}
Table~\ref{tab:attack_performance_cross_model} reports the cross-model attack results. The patch is trained on one surrogate and tested on others. Values are averaged across eight unseen datasets (excluding INRIA). \ours outperforms \tsea on most detectors across all three metrics, with a notable 21.43\% PASR gain when transferring from YOLO v2 to Faster R-CNN, demonstrating the effectiveness of our design.

\subsubsection{Cross-Dataset Attacks}
Table~\ref{tab:attack_performance_hr_transfer} reports the cross-dataset results on HR autonomous driving datasets. Values are averaged over seven unseen models (excluding the surrogate model).
We observe that \ours consistently outperforms \tsea on HR data, validating the positive effect of the proposed LCSL and PSPP in improving transferability under HR conditions. In particular, patches trained with \ours on different surrogate models achieve an average PASR of over 54\% on these datasets.

\subsubsection{Comparison with SOTA Competitors}
We report the cross-model transferability of partial competitors and our attack in Table~\ref{tab:attack_performance_competitors}. 
All patches are trained on YOLO v2 except for CAP and FDA, since only patches for YOLO v5 are available. 
We also provide the results for random patches as a reference. Intuitively, random noise results in the poorest attack performance. \ours achieves the best performance in nearly all cases. While some competitors come close to ours in terms of mAP, their PASR is significantly lower, which validates the overestimation issue of mAP mentioned before.

\begin{table*}[t]
\centering
\caption{Full attack performance compared with other competitors. \textbf{Bold} indicates the best performance.}
\label{tab:attack_performance_competitors}
\resizebox{0.98\linewidth}{!}{
\begin{tabular}{crrrrrrrrrrrrrrrrrrrrrrrrrrr}
\toprule
\multirow{2}{*}[-0.5ex]{Attack Method} & \multicolumn{3}{c}{YOLO v2} & \multicolumn{3}{c}{YOLO v3}  & \multicolumn{3}{c}{YOLO v3tiny} & \multicolumn{3}{c}{YOLO v4}  & \multicolumn{3}{c}{YOLO v4tiny} & \multicolumn{3}{c}{YOLO v5}  & \multicolumn{3}{c}{Faster-RCNN} & \multicolumn{3}{c}{SSD} \\
\cmidrule(r){2-4}\cmidrule(r){5-7}\cmidrule(r){8-10}\cmidrule(r){11-13}\cmidrule(r){14-16}\cmidrule(r){17-19}\cmidrule(r){20-22}\cmidrule(r){23-25}
& \footnotesize{PASR$\uparrow$} & \footnotesize{mAP$\downarrow$} & \footnotesize{ASR$\uparrow$} & \footnotesize{PASR$\uparrow$} & \footnotesize{mAP$\downarrow$} & \footnotesize{ASR$\uparrow$} & \footnotesize{PASR$\uparrow$} & \footnotesize{mAP$\downarrow$} & \footnotesize{ASR$\uparrow$} & \footnotesize{PASR$\uparrow$} & \footnotesize{mAP$\downarrow$} & \footnotesize{ASR$\uparrow$} & \footnotesize{PASR$\uparrow$} & \footnotesize{mAP$\downarrow$} & \footnotesize{ASR$\uparrow$} & \footnotesize{PASR$\uparrow$} & \footnotesize{mAP$\downarrow$} & \footnotesize{ASR$\uparrow$} & \footnotesize{PASR$\uparrow$} & \footnotesize{mAP$\downarrow$} & \footnotesize{ASR$\uparrow$} & \footnotesize{PASR$\uparrow$} & \footnotesize{mAP$\downarrow$} & \footnotesize{ASR$\uparrow$} \\
\midrule
Random 
& 16.42 & 81.74 & 17.13 & 16.99 & 87.61 & 12.09 & 20.81 & 81.45 & 18.36 & 18.22 & 86.81 & 13.09 & 23.09 & 80.27 & 19.56 & 22.14 & 83.15 & 16.77 & 23.07 & 82.50 & 17.04 & 18.21 & 84.00 & 15.85 \\
\midrule
AdvPatch
& 21.86 & 76.33 & 21.87 & 26.79 & 78.53 & 20.77 & 31.13 & 71.97 & 27.76 & 19.47 & 85.52 & 14.31 & 26.23 & 77.23 & 22.56 & 25.19 & 80.67 & 19.24 & 31.56 & 74.70 & 24.37 & 22.16 & 80.79 & 19.03 \\
\midrule
AdvTshirt
& 23.96 & 74.88 & 23.31 & 39.76 & 66.76 & 32.57 & 48.16 & 54.84 & 44.67 & 24.77 & 81.37 & 18.47 & 31.96 & 72.59 & 27.24 & 30.47 & 76.33 & 23.56 & 44.91 & 62.86 & 36.50 & 29.73 & 72.99 & 26.79 \\
\midrule
AdvCloak
& 24.24 & 74.52 & 23.79 & 32.23 & 73.60 & 25.64 & 39.81 & 63.26 & 36.30 & 20.36 & 84.71 & 15.13 & 27.34 & 76.26 & 23.56 & 26.54 & 79.41 & 20.49 & 34.10 & 72.24 & 26.99 & 23.32 & 79.26 & 20.56 \\
\midrule
AdvTexture
& 24.09 & 74.73 & 23.24 & 36.18 & 70.00 & 29.23 & 46.43 & 55.99 & 43.27 & 22.51 & 83.14 & 16.68 & 29.32 & 74.57 & 25.24 & 28.09 & 78.29 & 21.62 & 39.64 & 67.72 & 31.57 & 26.82 & 75.66 & 24.14 \\
\midrule
NAP
& 26.06 & 72.93 & 25.34 & 37.82 & 69.31 & 30.14 & 43.40 & 59.96 & 39.64 & 26.40 & 80.11 & 19.73 & 30.58 & 73.56 & 26.21 & 33.31 & 74.13 & 25.77 & 43.06 & 65.01 & 34.42 & 38.39 & 64.46 & 35.35 \\
\midrule
LAP
& 23.20 & 75.48 & 22.91 & 32.38 & 74.60 & 24.96 & 45.54 & 58.58 & 41.02 & 26.23 & 80.54 & 19.32 & 35.97 & 68.59 & 31.15 & 32.32 & 71.75 & 29.38 & 39.04 & 69.34 & 29.97 & 31.67 & 71.71 & 28.03 \\
\midrule
DAP
& 24.50 & 73.55 & 24.59 & 34.05 & 72.78 & 26.74 & 49.09 & 54.42 & 45.14 & 27.62 & 79.04 & 20.78 & 35.62 & 69.04 & 30.67 & 33.63 & 71.44 & 28.44 & 41.51 & 67.04 & 32.29 & 34.16 & 69.20 & 30.57 \\
\midrule
CAP
& 34.49 & 62.74 & 35.03 & 46.56 & 54.59 & 43.20 & 54.92 & 48.31 & 51.12 & 28.66 & 75.50 & 23.88 & 41.12 & 62.70 & 36.78 & \textbf{40.13} & 65.82 & 33.68 & 38.38 & 65.48 & 32.00 & 34.16 & 67.48 & 32.12 \\
\midrule
FDA
& 35.35 & 63.92 & 34.26 & 44.82 & 60.13 & 38.87 & 53.29 & 47.39 & 52.03 & 28.80 & 77.72 & 22.07 & 40.82 & 64.15 & 35.65 & 35.22 & 69.87 & 29.97 & 41.38 & 65.99 & 33.18 & 38.09 & 64.44 & 35.34 \\
\midrule
DePatch
& 35.06 & 60.67 & 36.35 & 39.93 & 55.94 & 39.31 & 54.23 & 40.62 & 57.50 & 22.43 & 79.37 & 18.93 & 40.92 & 59.51 & 38.71 & 31.05 & 73.50 & 25.66 & 34.86 & 67.08 & 29.27 & 25.67 & 75.65 & 23.69 \\
\midrule
MVPatch
& 32.39 & 65.66 & 32.13 & 36.57 & 68.82 & 30.00 & 48.29 & 54.66 & 44.75 & 28.78 & 77.06 & 22.52 & 42.20 & 61.65 & 37.78 & 33.35 & 69.09 & 30.57 & 38.67 & 68.06 & 30.62 & 34.23 & 68.92 & 30.73 \\
\midrule
AdvART
& 27.81 & 70.58 & 27.54 & 36.72 & 65.56 & 32.61 & 50.61 & 50.91 & 48.17 & 23.30 & 82.08 & 17.67 & 36.16 & 67.95 & 31.79 & 33.22 & 73.55 & 26.31 & 33.56 & 72.34 & 26.52 & 27.50 & 74.86 & 24.83 \\
\midrule
\tsea
& 67.18 & 20.71 & 73.26 & 36.16 & 32.22 & 52.28 & 50.95 & 31.71 & 60.52 & 23.16 & 67.17 & 26.41 & 44.11 & 44.98 & 48.95 & 31.67 & 63.96 & 31.54 & 26.03 & 53.39 & 32.41 & 35.18 & 57.28 & 38.86 \\
\midrule
\ours
& \textbf{71.28} & \textbf{12.83} & \textbf{79.95} & \textbf{50.24} & \textbf{16.45} & \textbf{67.40} & \textbf{55.53} & \textbf{25.84} & \textbf{66.22} & \textbf{31.20} & \textbf{55.13} & \textbf{34.18} & \textbf{50.52} & \textbf{30.79} & \textbf{59.64} & 35.74 & \textbf{56.35} & \textbf{36.42} & \textbf{47.46} & \textbf{27.00} & \textbf{52.55} & \textbf{49.50} & \textbf{35.64} & \textbf{57.60} \\

\bottomrule
\end{tabular}
}
\end{table*}

\subsection{Ablation Study}
\begin{figure}[t]
    \centering
    \includegraphics[width=0.98\linewidth]{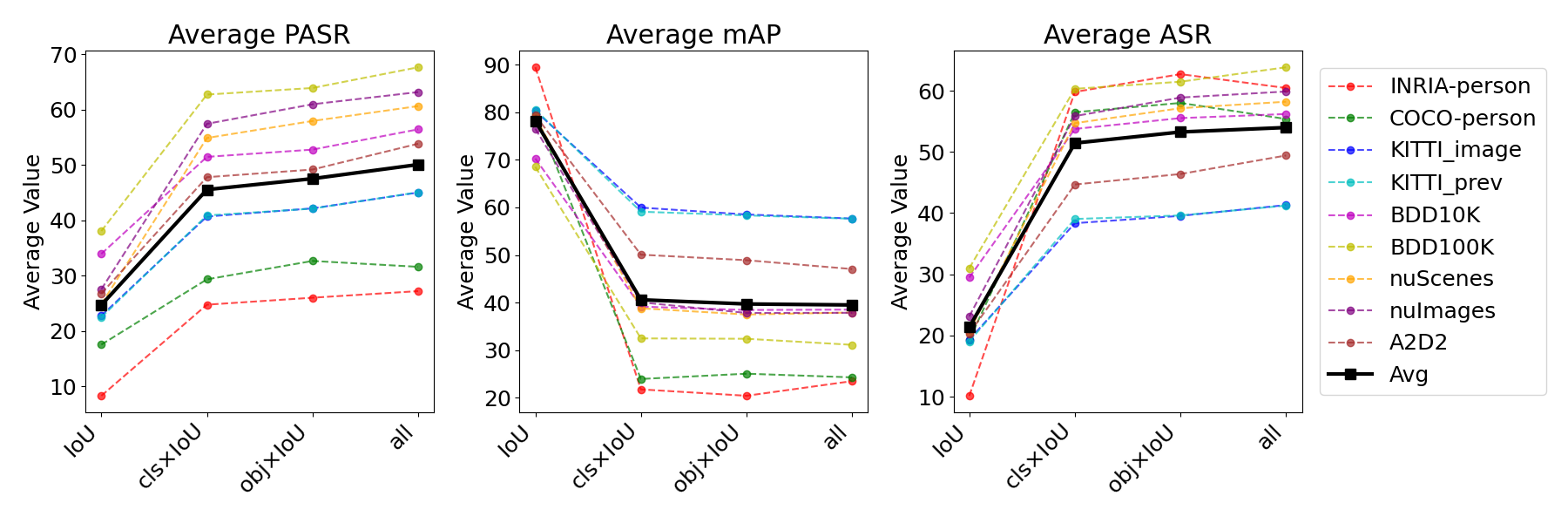}
    \caption{Ablation experiments on different loss designs.}
    \label{fig:ablation_method}
\end{figure}

We also conduct an ablation study to verify the effectiveness of our adversarial loss design. Specifically, we explore the following four loss variants with YOLO v3 being the surrogate model. 1) IoU: The top $T_k$ predictions are selected randomly and the maximum value of the IoU is considered the adversarial loss. 2) cls$\times$IoU: The top $T_k$ predictions are selected based on the classification score, and the maximum value of the product between the classification score and the IoU is taken. 3) obj$\times$IoU: Replace the classification score in cls$\times$IoU with the objectness score. 4) all: \ours. The results in Figure~\ref{fig:ablation_method} show that the combination of IoU and score information can indeed improve the average transferability. Besides, \ours performs the best in all metrics, while resting solely on IoU does not benefit the attack due to the lack of score guidance.

\subsection{Countermeasures}
We further evaluate \ours against two SOTA patch defenses, PAD~\cite{jing2024pad} and NAPGuard~\cite{wu2024napguard}, using patches trained on YOLO v2.
Table~\ref{tab:defense} reports the defense performance of PAD~\cite{jing2024pad} and NAPGuard~\cite{wu2024napguard}. 
We assess their effectiveness on images from five datasets, using patches trained on YOLO v2. PAD is a patch removal method that detects the patch position and replaces the patch with black pixels. We evaluate the attack performance on images before and after PAD.
NAPGuard is a detector that identifies the patch position along with a confidence score; therefore, we assess its performance using AP@0.5 and mAP@0.5:0.95. The results show that both defenses perform well on INRIA, since they are trained on INRIA. However, on HR datasets, PAD often fails to detect the patch or can only identify a small part of it, resulting in minimal changes to the patch region. These observations explain why the PASR after PAD is comparable to or even higher than before PAD. Additionally, PAD tends to misidentify unrelated regions as the patch, for example, mistakenly identifying background buildings behind pedestrians as patches. As for NAPGuard, the AP of NAPGuard drops significantly on HR datasets. We find that NAPGuard often predicts smaller bounding boxes or even fails to detect the patch, leading to a 50\% drop in AP@0.5 and an 80\% drop in mAP@0.5:0.95 compared to the results on INRIA. Overall, these findings suggest that PAD and NAPGuard perform poorly when dealing with small objects, and cannot effectively resist our attack.

\begin{table}[t]
\centering
\caption{Defense performance.}
\label{tab:defense}
\resizebox{0.98\linewidth}{!}{
\begin{tabular}{llccccc}
\toprule
Defense & Metric & INRIA & KITTI-i & bdd100k & nuImages & A2D2 \\
\midrule
\multirow{2}{*}{PAD} & PASR before PAD & 23.25 & 38.37 & 60.04 & 54.44 & 43.12 \\
& PASR after PAD & 18.02 & 50.68 & 57.32 & 48.05 & 48.40 \\
\midrule
\multirow{2}{*}{NAPGuard} & AP@0.5 & 99.46 & 58.15 & 50.91 & 42.92 & 49.73 \\
& mAP@0.5:0.95 & 76.64 & 20.95 & 18.93 & 13.90 & 18.87 \\
\bottomrule
\end{tabular}
}
\end{table}

\subsection{Physical Attacks}
To verify that our adversarial patch can be deployed in real-world driving scenarios, we conduct a simple physical attack. 
The experimental settings are as follows:
\begin{itemize}
    \item Device: iPhone 15 pro, 720p@30fps, inside the vehicle.
    \item Model used: same object detectors in digital experiments.
    \item Vehicle speed during recording: $\sim$7 km/h.
    \item Patch printing: the patch is printed on a piece of A4 paper and attached to front chest of the pedestrian.
    \item Recording distance to pedestrian: 20 m - 0 m.
    \item Recording image resolution: 1,280×720.
    \item Pedestrian pose and motion: standing still, facing camera.
\end{itemize}

We clarify that it is difficult to strictly control all environmental conditions in physical-world testing. Variations in background, weather, lighting, and the precise relative positioning between the vehicle and pedestrian can introduce differences between baseline and attack recordings, although we have tried our best to minimize such impact. We ensure that the vehicle, camera device, and pedestrian subject remains identical across baseline and attack sequences. Moreover, the test is conducted in the same road and time window to minimize external variations.

\begin{figure}[t]
    \centering
    \includegraphics[width=0.95\linewidth]{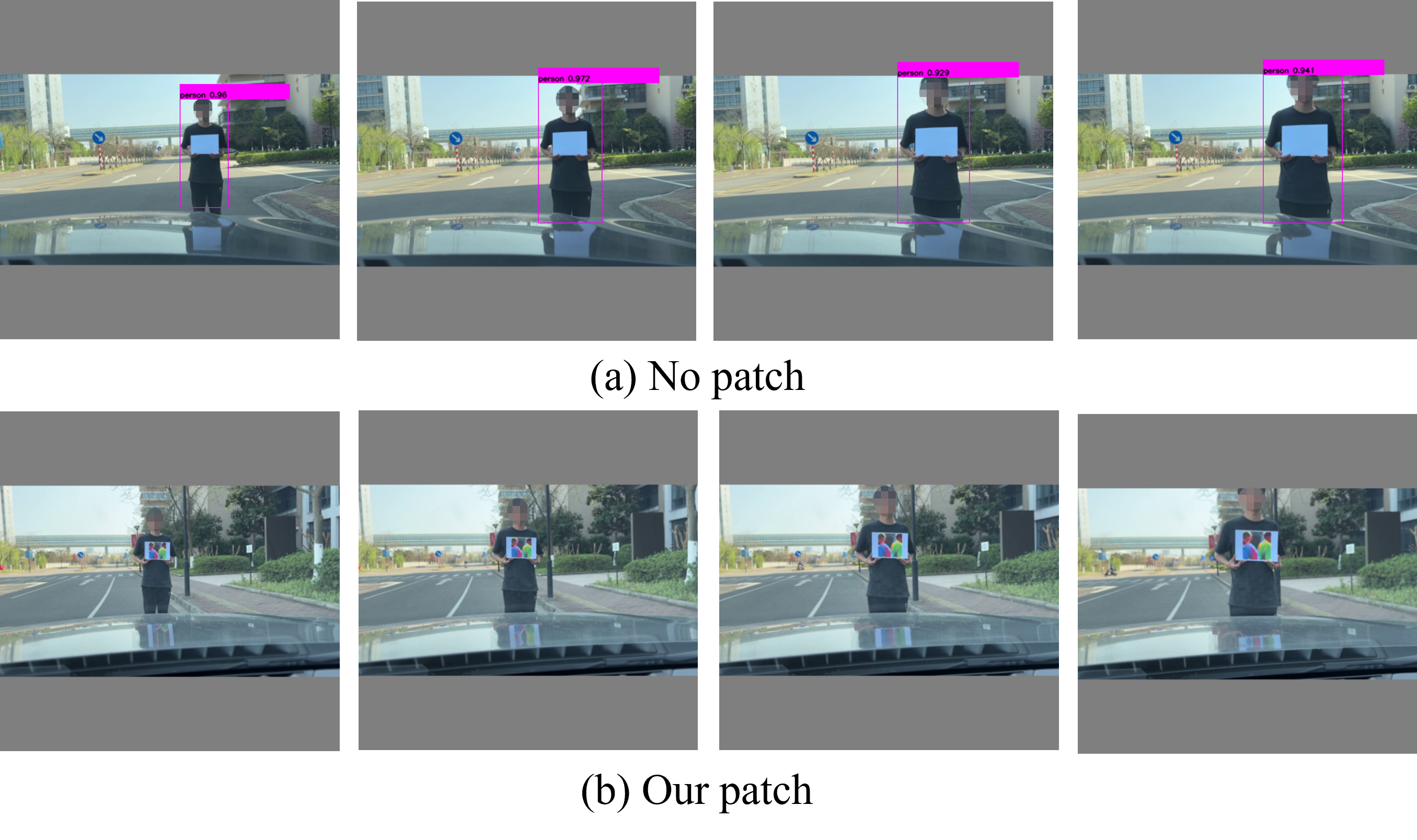}
    \caption{Visualizations for physical attacks.}
    \label{fig:physical}
\end{figure}

As shown in Figure~\ref{fig:physical}, we capture videos inside a moving car as it continuously approaches a person holding the printed patch, until the car collided with the person. Note that we only focus on close-range scenarios, as successful attacks at longer distances are unlikely to cause significant safety risks.
We observe that the detector reliably detects the person when holding a blank paper (confidence$>$0.9), but fails completely when the adversarial patch is present.
More importantly, from the moment the distance between the car and the person falls below 4 meters until the collision, YOLO v4-tiny fails to detect the person for a continuous period exceeding 2 seconds. This demonstrates the strong practical effectiveness and highlights the potential impact of our patch in real-world scenarios.

We conduct additional physical-world experiments using adversarial patches generated by several competitors, under the same deployment and recording setup. Specifically, we evaluate the longest continuous duration (both in frames and seconds) during which a pedestrian is completely undetected in the video sequence. This metric better captures the practical impact of a patch in safety-critical scenarios. As shown in Table~\ref{tab:quantitative_physical}, our proposed \ours achieves significantly longer invisibility durations than other competitors, demonstrating its superior effectiveness in real-world conditions.

\begin{table}[!t]
\centering
\caption{Quantitative results on physical attacks.}
\label{tab:quantitative_physical}
\begin{tabular}{lcc}
\toprule
Method & Duration (frames) & Duration (seconds) \\
\midrule
AdvPatch & 12 & 0.40 \\
NAP  & 5 & 0.17 \\
\tsea & 9 & 0.30 \\
\ours & 69 & 2.30 \\
\bottomrule
\end{tabular}
\end{table}

\section{Conclusion}
This paper focuses on addressing the challenges of transferring adversarial patch attacks to HR autonomous driving datasets. We first analyze the overestimation issues present in existing mAP-based evaluation metrics and then identify the limitations of current adversarial loss designs, which often overlook the role of IoU. To solve these limitations, we propose PASR, a more practical and safety-aligned metric that can accurately reflect the true impact on the detector. Furthermore, we introduce the LCSL that effectively combines IoU with detection scores. In addition, we apply the PSPP during data preprocessing to improve the transferability of patches trained on LR datasets to HR data. Extensive experiments demonstrate that our proposed \ours achieves significantly stronger cross-model and cross-dataset transferability compared with existing attacks, makes the patch harder to be detected, and is capable of causing long-term detection failures in physical-world autonomous driving scenarios. 

\noindent\textbf{Limitations and Future Work.} Following T-SEA, we prioritize transferability over patch semantics, although it has been studied in some prior works. We leave exploring the trade-off between transferability and naturalness for future work. In addition, we will explore potential defenses against such attacks and further improve the proposed IoU-based loss for transformer-based and multimodal detectors.

\bibliographystyle{IEEEtran}
\bibliography{ref}

\onecolumn
\newpage
\twocolumn

\section*{Appendix}
\setcounter{section}{0}
\renewcommand{\thesection}{\Alph{section}}

\section{Justification for Threat Model}
In our threat model, we assume that the attacker can only use LR data to train the adversarial patch, rather than directly training on HR data. Although it is indeed intuitive that the latter assumption could yield stronger attack performance on HR datasets, such an approach suffers from several significant limitations that undermine its practicality and transferability in real-world scenarios.

1) Computational cost and memory constraints: HR training dramatically increases GPU memory consumption and slows down optimization. Taking the A2D2 dataset as an example, a single 1,920×1,208 image in A2D2 is approximately 46 times larger in pixel count than a 224×224 image, leading to exponential growth in training time and severely restricted batch sizes. This makes training infeasible for large-scale patch optimization with object detectors.

2) Overfitting and weak transferability: HR datasets often contain highly uniform scenes with limited diversity. Patches trained on one such dataset may overfit to dataset-specific spatial patterns and fail to transfer to other domains/datasets. Moreover, in many autonomous driving datasets, pedestrians are small in scale, making the learning ability weak and the patch less effective. Therefore, these issues severely hinder transferability.

3) Dataset availability and annotation quality: Large-scale, HR datasets with dense, high-quality pedestrian annotations are limited. Most existing HR datasets (\eg bdd100k, A2D2) were not designed with patch training in mind, and may contain inconsistencies in labeling, pedestrian scale, and image quality. This makes patch optimization over such data difficult and unreliable.

Considering these limitations, we adopt a practical and generalizable setting: training on a LR, general person detection dataset such as INRIA, and evaluating on multiple HR autonomous driving datasets. This allows us to keep computational costs low while demonstrating strong cross-dataset transferability.
Therefore, while training on HR data may improve performance under controlled conditions, it is not a practical or scalable solution for real-world attack deployment. Our method provides a more feasible, lightweight, and transferable alternative.

\section{Justification for PASR}
We clarify that PASR does not evaluate per-person attack success, but whether an image contains a safety risk, \ie whether any pedestrian remains entirely undetected (Equation~\ref{equ:image_level}). In autonomous driving, missing even one critical pedestrian (especially one near or in the vehicle's path) may prevent the vehicle from slowing or stopping, regardless of how many others are detected. The decision logic depends on whether any critical pedestrian is missed, not on counts or averages.
This reveals a key limitation of mAP: it averages over all detections and may remain high even when the most safety-critical target is missed, failing to reflect real safety risk. In contrast, PASR directly captures this by evaluating whether any pedestrian remains fully undetected, better aligning with safety-critical risks.

Regarding cases of low IoU detections, we clarify that small but non-zero IoU may still offer weak cues and inform the system of a potential hazard, and will trigger cautious behavior. However, when IoU=0 (no overlap), the system may remain unaware of the pedestrian and proceed unsafely.

\noindent\textbf{Regulatory references.} Our safety assumption is also grounded in widely adopted regulatory standards and real-world system validation protocols, especially for autonomous driving systems where pedestrian detection is a critical perception task. Multiple regulations and safety frameworks explicitly or implicitly follow a zero-tolerance policy toward pedestrian detection failures, meaning that the failure to detect even a single pedestrian in a test scenario is considered unacceptable. For example, the U.S. federal regulation, FMVSS No.127 (NHTSA), mandates automatic emergency braking (AEB) systems for light vehicles to detect and respond to pedestrians. The regulation explicitly states: ``FMVSS127 requires a 100\% pass rate for a mandated test, leaving no room for error''
\footnote{\url{https://www.abdynamics.com/what-does-the-upcoming-fmvss-127-aeb-protocol-mean-for-automotive-testing/}}.
Similar expression can also be found in Euro NCAP AEB Pedestrian Protocol \footnote{\url{https://www.euroncap.com}} and UNECE Regulation No. 127 \footnote{\url{https://unece.org/}}.
These regulations emphasize fail-operational perception and safety-critical completeness. As such, our evaluation metric PASR is intentionally designed to reflect any failure to detect a single pedestrian as a safety risk, in alignment with both regulatory expectations and real-world safety demands. 

\section{More Experimental Results}
\begin{figure}[t]
    \centering
    \includegraphics[width=0.98\linewidth]{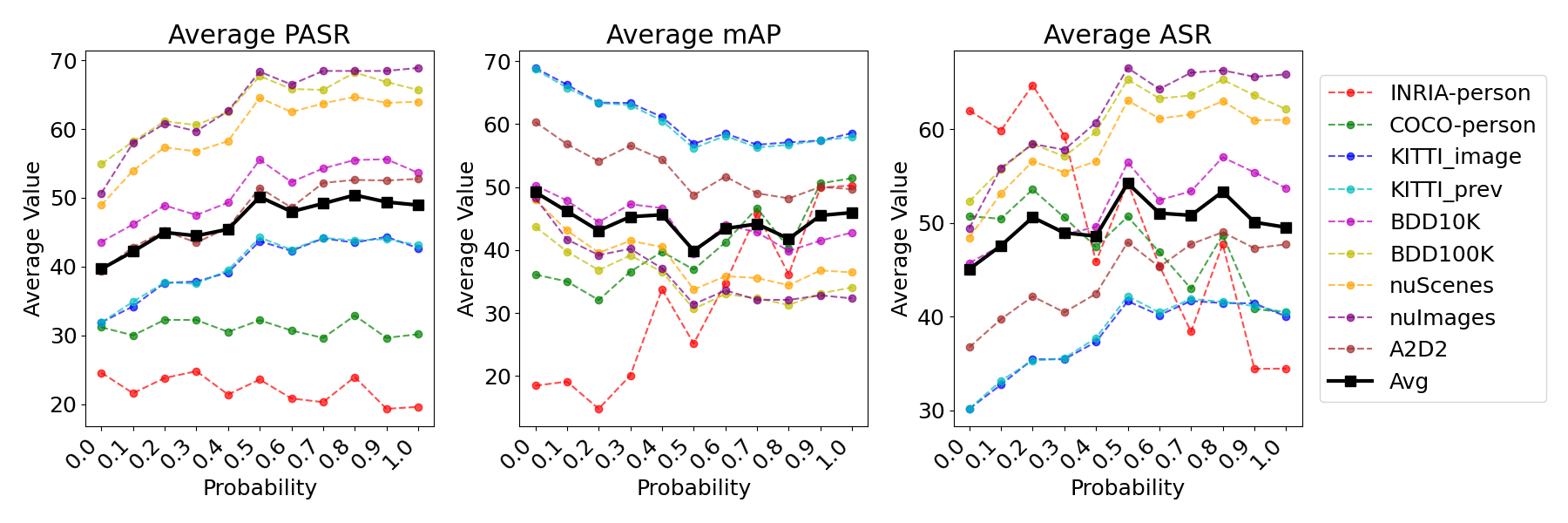}
    \caption{Hyperparameter selection for probability.}
    \label{fig:ablation_p}
\end{figure}

\noindent\textbf{Hyperparameter Search.}
Recall that we introduce a data preprocessing step to improve the transferability to HR datasets. Consequently, the preprocessing probability $p_{hr}$ becomes a significant hyperparameter affecting the overall performance. To determine the optimal value, we perform a grid search by varying $p_{hr}$ from 0.0 to 1.0. The patch is trained on Faster-RCNN. 
As shown in Figure~\ref{fig:ablation_p}, transferability generally improves as $p_{hr}$ increases, but fluctuates when $p_{hr}$ is close to 1.0. This is attributed to the reduced effectiveness of the LR dataset (COCO) and the overfitting to fine-grained details in HR images. Based on these observations, we set $p_{hr}=0.5$ as it achieves the best overall transferability.

\noindent\textbf{Cross-dataset Attack Analyses.}
We provide the results for cross-dataset transferability on LR datasets in Table~\ref{tab:attack_performance_lr_transfer}. 
The data in the table represents the average values across seven unseen models. \ours can achieve comparable performance compared with T-SEA. Due to the incorporation of the Probabilistic Scale-Preserving Padding during the training process of our patch, \ours is slightly outperformed by T-SEA in a few cases, as T-SEA is trained solely on LR datasets and thus achieves better results under such conditions. The results on the two LR datasets are provided just for reference, as this work primarily focuses on improving transferability on HR datasets, which are more representative of practical real-world scenarios.
It is also worth noting that the table reveals and highlights the limitations of mAP-based metrics, as discussed in main text. In some cases, the attack method with a lower mAP exhibits a lower PASR, indicating that mAP cannot accurately reflect the true attack effectiveness. 

\begin{table}[!t]  
\centering
\caption{{Attack performance of cross-dataset transferability (LR data). Imp. denotes improvement. \textcolor{blue}{Blue} indicates better performance of our attack; \textcolor{red}{red} denotes worse.}}
\label{tab:attack_performance_lr_transfer}
\resizebox{0.8\linewidth}{!}{
\begin{tabular}{ccrrrrrrrrrrrrrrrrrrrrrrrrrrrrrr}
\toprule
\multirow{2}{*}[-0.5ex]{\diagbox{Train}{Test}} & 
\multirow{2}{*}[-0.5ex]{Method} & \multicolumn{3}{c}{INRIA} & \multicolumn{3}{c}{COCO} \\
\cmidrule(r){3-5}\cmidrule(r){6-8}
& & \footnotesize{PASR$\uparrow$} & \footnotesize{mAP$\downarrow$} & \footnotesize{ASR$\uparrow$} & \footnotesize{PASR$\uparrow$} & \footnotesize{mAP$\downarrow$} & \footnotesize{ASR$\uparrow$} \\
\midrule
\multirow{3}{*}{YOLO v2} &
\tsea
& 24.51 & 31.61 & 53.42 & 30.99 & 35.77 & 52.23 \\
& \ours
& 27.54 & 21.02 & 61.35 & 31.99 & 20.46 & 58.89 \\
& Imp.
& \textcolor{blue}{3.03} & \textcolor{blue}{10.59} & \textcolor{blue}{7.93} & \textcolor{blue}{1.00} & \textcolor{blue}{15.31} & \textcolor{blue}{6.66} \\

\midrule
\multirow{3}{*}{YOLO v3} &
\tsea
& 26.29 & 53.93 & 39.50 & 38.88 & 47.77 & 47.02 \\
& \ours
& 27.23 & 23.48 & 60.45 & 31.62 & 24.27 & 55.38 \\
& Imp.
& \textcolor{blue}{0.94} & \textcolor{blue}{30.45} & \textcolor{blue}{20.95} & \textcolor{red}{-7.26} & \textcolor{blue}{23.50} & \textcolor{blue}{8.36} \\

\midrule
\multirow{3}{*}{YOLO v3tiny} &
\tsea
& 21.13 & 59.60 & 33.28 & 26.48 & 47.82 & 39.04 \\
& \ours
& 23.63 & 51.85 & 40.55 & 29.84 & 41.97 & 48.36 \\
& Imp.
& \textcolor{blue}{2.50} & \textcolor{blue}{7.75} & \textcolor{blue}{7.27} & \textcolor{blue}{3.36} & \textcolor{blue}{5.85} & \textcolor{blue}{9.32} \\

\midrule
\multirow{3}{*}{YOLO v4} &
\tsea
& 23.27 & 24.57 & 41.42 & 30.50 & 30.63 & 43.16 \\
& \ours
& 17.36 & 22.44 & 54.59 & 26.83 & 24.26 & 53.97 \\
& Imp.
& \textcolor{red}{-5.91} & \textcolor{blue}{2.13} & \textcolor{blue}{13.17} & \textcolor{red}{-3.67} & \textcolor{blue}{6.37} & \textcolor{blue}{10.81} \\

\midrule
\multirow{3}{*}{YOLO v4tiny} &
\tsea
& 22.64 & 57.71 & 34.98 & 31.04 & 45.52 & 44.71 \\
& \ours
& 28.11 & 56.62 & 39.40 & 36.73 & 49.96 & 46.32 \\
& Imp.
& \textcolor{blue}{5.47} & \textcolor{blue}{1.09} & \textcolor{blue}{4.42} & \textcolor{blue}{5.69} & \textcolor{red}{-4.44} & \textcolor{blue}{1.61} \\

\midrule
\multirow{3}{*}{YOLO v5} &
\tsea
& 28.41 & 55.73 & 40.60 & 36.77 & 51.52 & 42.77 \\
& \ours
& 21.65 & 34.16 & 48.61 & 31.08 & 41.40 & 47.24 \\
& Imp.
& \textcolor{red}{-6.76} & \textcolor{blue}{21.57} & \textcolor{blue}{8.01} & \textcolor{red}{-5.69} & \textcolor{blue}{10.12} & \textcolor{blue}{4.47} \\

\midrule
\multirow{3}{*}{Faster-RCNN} &
\tsea
& 17.59 & 66.64 & 28.51 & 29.29 & 58.07 & 37.57 \\
& \ours
& 23.58 & 25.13 & 54.32 & 32.20 & 36.89 & 50.67 \\
& Imp.
& \textcolor{blue}{5.99} & \textcolor{blue}{41.51} & \textcolor{blue}{25.81} & \textcolor{blue}{2.91} & \textcolor{blue}{21.18} & \textcolor{blue}{13.10} \\

\midrule
\multirow{3}{*}{SSD} &
\tsea
& 23.45 & 64.62 & 33.06 & 34.93 & 54.57 & 38.60 \\
& \ours
& 31.09 & 62.23 & 36.42 & 41.29 & 52.08 & 46.09 \\
& Imp.
& \textcolor{blue}{7.64} & \textcolor{blue}{2.39} & \textcolor{blue}{3.36} & \textcolor{blue}{6.36} & \textcolor{blue}{2.49} & \textcolor{blue}{7.49} \\

\bottomrule
\end{tabular}
}
\end{table}

\noindent\textbf{Cross-model Attack on New Models.}
In our experiments, we choose YOLO series, Faster-RCNN and SSD as detectors to ensure fair comparison with prior works, (\eg T-SEA, MVPatch, and NAP), which also use these models in their evaluations. That said, evaluation on newer detectors is also important. Therefore, we additionally evaluate our method on three more recent object detectors, RT-DETR2~\cite{lv2024Rt-detrv2}, YOLOv8~\cite{yolov8_2023ultralytics} and YOLO11~\cite{yolo11_2024ultralytics}. Results in Table~\ref{tab:attack_performance_cross_model_new_models} show that our trained patches remain effective across these models. 

\begin{table}[!t]  
\centering
\caption{Attack performance of cross-model transferability (new models). Imp. denotes improvement. \textcolor{blue}{Blue} indicates better performance of our attack; \textcolor{red}{red} denotes worse. }
\label{tab:attack_performance_cross_model_new_models}
\resizebox{0.97\linewidth}{!}{
\begin{tabular}{ccrrrrrrrrrrrrrrrrrrrrrrrrrrr}
\toprule
\multirow{2}{*}[-0.5ex]{\diagbox{Train}{Test}} & 
\multirow{2}{*}[-0.5ex]{Method} & \multicolumn{3}{c}{RT-DETRv2} & \multicolumn{3}{c}{YOLO v8}  & \multicolumn{3}{c}{YOLO 11} \\
\cmidrule(r){3-5}\cmidrule(r){6-8}\cmidrule(r){9-11}
& & \footnotesize{PASR$\uparrow$} & \footnotesize{mAP$\downarrow$} & \footnotesize{ASR$\uparrow$} & \footnotesize{PASR$\uparrow$} & \footnotesize{mAP$\downarrow$} & \footnotesize{ASR$\uparrow$} & \footnotesize{PASR$\uparrow$} & \footnotesize{mAP$\downarrow$} & \footnotesize{ASR$\uparrow$} &  \\
\midrule
\multirow{3}{*}{YOLO v2} & \tsea  
& 33.35 & 49.91 & 36.91 & 32.35 & 61.81 & 34.53 & 40.47 & 54.74 & 41.61 \\
& \ours 
& 42.16 & 29.85 & 50.11 & 37.78 & 53.08 & 39.94 & 45.12 & 44.46 & 47.04 \\
& Imp.
& \textcolor{blue}{8.81} & \textcolor{blue}{-20.06} & \textcolor{blue}{13.20} & \textcolor{blue}{5.43} & \textcolor{blue}{-8.73} & \textcolor{blue}{5.41} & \textcolor{blue}{4.65} & \textcolor{blue}{-10.28} & \textcolor{blue}{5.43} \\

\midrule
\multirow{3}{*}{YOLO v3} & \tsea 
& 43.63 & 58.40 & 39.80 & 35.17 & 68.12 & 31.27 & 47.76 & 55.64 & 43.74 \\
& \ours 
& 46.12 & 45.70 & 44.27 & 41.03 & 56.98 & 39.07 & 51.76 & 41.92 & 51.86 \\
& Imp.
& \textcolor{blue}{2.49} & \textcolor{blue}{-12.70} & \textcolor{blue}{4.47} & \textcolor{blue}{5.86} & \textcolor{blue}{-11.14} & \textcolor{blue}{7.80} & \textcolor{blue}{4.00} & \textcolor{blue}{-13.72} & \textcolor{blue}{8.12} \\

\midrule
\multirow{3}{*}{YOLO v3tiny} & \tsea 
& 42.86 & 59.70 & 37.92 & 34.48 & 68.42 & 30.69 & 45.02 & 58.59 & 40.32 \\
& \ours 
& 46.20 & 56.47 & 41.08 & 38.44 & 64.31 & 34.67 & 49.73 & 52.95 & 45.78 \\
& Imp.
& \textcolor{blue}{3.34} & \textcolor{blue}{-3.23} & \textcolor{blue}{3.16} & \textcolor{blue}{3.96} & \textcolor{blue}{-4.11} & \textcolor{blue}{3.98} & \textcolor{blue}{4.71} & \textcolor{blue}{-5.64} & \textcolor{blue}{5.46} \\

\midrule
\multirow{3}{*}{YOLO v4} & \tsea 
& 32.06 & 54.04 & 33.56 & 32.00 & 62.29 & 33.91 & 39.73 & 53.95 & 41.42 \\
& \ours 
& 33.90 & 48.36 & 35.37 & 34.54 & 58.41 & 35.39 & 39.44 & 50.63 & 41.24 \\
& Imp.
& \textcolor{blue}{1.84} & \textcolor{blue}{-5.68} & \textcolor{blue}{1.81} & \textcolor{blue}{2.54} & \textcolor{blue}{-3.88} & \textcolor{blue}{1.48} & \textcolor{red}{-0.29} & \textcolor{blue}{-3.32} & \textcolor{red}{-0.18} \\

\midrule
\multirow{3}{*}{YOLO v4tiny} & \tsea 
& 42.00 & 57.03 & 38.82 & 29.72 & 71.99 & 27.05 & 43.02 & 58.57 & 39.20 \\
& \ours 
& 44.30 & 59.94 & 39.17 & 37.38 & 67.11 & 32.53 & 47.55 & 57.14 & 42.42 \\
& Imp.
& \textcolor{blue}{2.30} & \textcolor{red}{2.91} & \textcolor{blue}{0.35} & \textcolor{blue}{7.66} & \textcolor{blue}{-4.88} & \textcolor{blue}{5.48} & \textcolor{blue}{4.53} & \textcolor{blue}{-1.43} & \textcolor{blue}{3.22} \\

\midrule
\multirow{3}{*}{YOLO v5} & \tsea 
& 37.42 & 64.55 & 33.11 & 33.55 & 69.25 & 29.86 & 43.79 & 58.53 & 40.21 \\
& \ours 
& 41.32 & 54.15 & 40.18 & 35.50 & 61.06 & 35.37 & 46.74 & 51.85 & 44.91 \\
& Imp.
& \textcolor{blue}{3.90} & \textcolor{blue}{-10.40} & \textcolor{blue}{7.07} & \textcolor{blue}{1.95} & \textcolor{blue}{-8.19} & \textcolor{blue}{5.51} & \textcolor{blue}{2.95} & \textcolor{blue}{-6.68} & \textcolor{blue}{4.70} \\

\midrule
\multirow{3}{*}{Faster-RCNN} & \tsea 
& 40.19 & 60.83 & 36.09 & 37.68 & 65.19 & 33.76 & 44.13 & 57.76 & 40.69 \\
& \ours 
& 45.79 & 45.97 & 45.10 & 39.67 & 58.04 & 38.36 & 48.89 & 47.71 & 47.88 \\
& Imp.
& \textcolor{blue}{5.60} & \textcolor{blue}{-14.86} & \textcolor{blue}{9.01} & \textcolor{blue}{1.99} & \textcolor{blue}{-7.15} & \textcolor{blue}{4.60} & \textcolor{blue}{4.76} & \textcolor{blue}{-10.05} & \textcolor{blue}{7.19} \\

\midrule
\multirow{3}{*}{SSD} & \tsea 
& 41.47 & 61.92 & 37.60 & 38.55 & 64.53 & 34.95 & 47.00 & 56.37 & 42.83 \\
& \ours 
& 45.97 & 59.86 & 39.61 & 43.18 & 62.27 & 37.60 & 50.47 & 55.07 & 44.73 \\
& Imp.
& \textcolor{blue}{4.50} & \textcolor{blue}{-2.06} & \textcolor{blue}{2.01} & \textcolor{blue}{4.63} & \textcolor{blue}{-2.26} & \textcolor{blue}{2.65} & \textcolor{blue}{3.47} & \textcolor{blue}{-1.30} & \textcolor{blue}{1.90} \\

\bottomrule
\end{tabular}
}
\end{table}

\noindent\textbf{Failure Case Analysis.}
In some cases, we observe that the detector still outputs the person but with multiple fragmented boxes, such as one on the head and another covering the patch and lower body. A plausible explanation is that during training, the patch may have been optimized to mimic features resembling a human head, which, while reducing IoU with the original GT box, still triggers partial detections. These cases are consistent with the goal of lowering IoU but illustrate the limits of patch-based disappearance. 
 
\section{Theoretical Analysis of Generalization Bound}
In this section, we provide a theoretical analysis to characterize the generalization ability of the proposed adversarial patch training framework. While our patch attack involves maximization over confidence scores and IoU, we demonstrate that its capacity remains well-controlled in terms of empirical Rademacher complexity. This guarantees that the learned patch is not overly prone to overfitting, even under the highly structured optimization objective. 

Recall that our adversarial loss is expressed as follows.
\begin{equation}
\mathcal{L}_{\text{adv}} = \frac{1}{N_b} \sum\nolimits_{i=1}^{N_b} \max_{(\bar{\mathbf{b}}_j, \bar{o}_j, \bar{\mathbf{s}}_j) \in \bar{\mathbf{O}}_{T_k}} \left[ \bar{o}_j \cdot \bar{s}_j^0 \cdot \text{IoU}\left( \bar{\mathbf{b}}_j, \mathbf{b}_{\max}^{\text{GT}} \right) \right].
\end{equation}
During training, we optimize this loss over mini-batches, where each mini-batch loss is computed as the average over individual image losses within the batch. However, for generalization analysis based on Rademacher complexity, we focus on the population-level function class that maps individual input samples to scalar loss values. Since Rademacher complexity is defined at the level of individual samples, the averaging over mini-batches in the training process does not affect the complexity bound derivation.

The empirical Rademacher complexity of the loss function class $\mathcal{F}$ associated with $\mathcal{L}_{\text{adv}}$ is defined as:
\begin{equation}
\widehat{\mathfrak{R}}_X(\mathcal{F}) = \mathbb{E}_{\boldsymbol{\sigma}} \left[ \sup_{\mathcal{L}_{\mathrm{adv}} \in \mathcal{F}} \frac{1}{|X|} \sum_{i=1}^{|X|} \sigma_i \, \mathcal{L}_{\mathrm{adv}}(x_i) \right],
\end{equation}
where \( \sigma_i \in \{-1, +1\} \) are independent Rademacher random variables, $f$ denotes the object detector, $X$ denotes the training images.

Let $h_j = \bar{o}_j \cdot \bar{s}_j^0 \cdot \mathrm{IoU}\left( \bar{\mathbf{b}}_j, \mathbf{b}_{\max}^{\mathrm{GT}} \right)$, we rewrite $\mathcal{L}_{\text{adv}}$ as:
\begin{equation}
\mathcal{L}_{\mathrm{adv}}(\mathbf{x}) = \max_{j \in \{1, \ldots, {T_k}\}} h_j(\mathbf{x}),
\end{equation}

By applying Massart's finite-class maximal inequality~\cite{shalev2014understanding}, we obtain:

\begin{equation}
\widehat{\mathfrak{R}}_X(\mathcal{F}) \leq \max_{j \in \{1, \ldots, T_k\}} \widehat{\mathfrak{R}}_X(\mathcal{H}_j) + \sqrt{\frac{2 \log T_k}{|X|}},
\end{equation}
where \( \mathcal{H}_j \) is the function class corresponding to \( h_j \). Each \( h_j \) is a product of three functions:
\begin{equation}
h_j(\mathbf{x}) = f_j^{\mathrm{obj}}(\mathbf{x}) \cdot f_j^{\mathrm{cls}}(\mathbf{x}) \cdot f_j^{\mathrm{IoU}}(\mathbf{x}),
\end{equation}
where $f_j^{\mathrm{obj}}(\mathbf{x}) = \bar{o}_j$, $\quad f_j^{\mathrm{cls}}(\mathbf{x}) = \bar{s}_j^0$, $\quad f_j^{\mathrm{IoU}}(\mathbf{x}) = \mathrm{IoU}\left( \bar{\mathbf{b}}_j, \mathbf{b}_{\max}^{\mathrm{GT}} \right)$.

For \( f_j^{\mathrm{obj}} \) and \( f_j^{\mathrm{cls}} \), we assume they are Lipschitz continuous with respect to the input, with Lipschitz constants \( \alpha_{\mathrm{obj}} \) and \( \alpha_{\mathrm{cls}} \), such that:

\begin{equation}
\left| f_j^{\mathrm{obj}}(\mathbf{x}) - f_j^{\mathrm{obj}}(\mathbf{x}_{adv}) \right| \leq \alpha_{\mathrm{obj}} \cdot \| x - \mathbf{x}_{adv} \|_2,
\end{equation}

\begin{equation}
\left| f_j^{\mathrm{cls}}(\mathbf{x}) - f_j^{\mathrm{cls}}(\mathbf{x}_{adv}) \right| \leq \alpha_{\mathrm{cls}} \cdot \| x - \mathbf{x}_{adv} \|_2.
\end{equation}

The standard IoU function is not globally Lipschitz due to discontinuities when predicted and GT boxes have no overlaps. To ensure smoothness for theoretical analysis, we adopt a smoothed IoU formulation:

\begin{equation}
\widetilde{\mathrm{IoU}}\left( \bar{\mathbf{b}}_j, \mathbf{b}_{\max}^{\mathrm{GT}} \right) = \frac{|\bar{\mathbf{b}}_j \cap \mathbf{b}_{\max}^{\mathrm{GT}}| + \epsilon}{|\bar{\mathbf{b}}_j \cup \mathbf{b}_{\max}^{\mathrm{GT}}| + \epsilon},
\end{equation}
where \( \epsilon > 0 \) is a small constant that ensures differentiability even for disjoint boxes. With this smoothing, \( f_j^{\mathrm{IoU}}(\mathbf{x}) = \widetilde{\mathrm{IoU}}\left( \bar{\mathbf{b}}_j, \mathbf{b}_{\max}^{\mathrm{GT}} \right) \) is Lipschitz continuous with constant \( \alpha_{\mathrm{IoU}} \), satisfying:

\begin{equation}
\left| f_j^{\mathrm{IoU}}(\mathbf{x}) - f_j^{\mathrm{IoU}}(\mathbf{x}_{adv}) \right| \leq \alpha_{\mathrm{IoU}} \cdot \| x - \mathbf{x}_{adv} \|_2.
\end{equation}

For any image \( x \), after pasting the patch onto the detected persons, the effective perturbation norm remains bounded due to the finite image resolution and finite number of pasted instances. Therefore, the perturbation magnitude induced by applying the patch is upper bounded by:

\begin{equation}
B := \sup_{x \in X} \| \mathbf{x}_{adv} - \mathbf{x} \|_2,
\end{equation}

By applying the Rademacher inequality, we obtain:
\begin{equation}
\widehat{\mathfrak{R}}_{X}(h_j) \leq \frac{(\alpha_{\mathrm{obj}} + \alpha_{\mathrm{cls}} + \alpha_{\mathrm{IoU}}) B}{\sqrt{|X|}}.
\end{equation}

Finally, substituting into the aggregation bound yields the final complexity bound:

\begin{equation}
\widehat{\mathfrak{R}}_{X}(\mathcal{F}) \leq \frac{(\alpha_{\mathrm{obj}} + \alpha_{\mathrm{cls}} + \alpha_{\mathrm{IoU}}) B}{\sqrt{|X|}} + \sqrt{\frac{2 \log T_k}{|X|}}.
\end{equation}

This result demonstrates that the overall generalization complexity remains well-controlled. On the other hand, the multiplicative structure in our loss design is better aligned with our proposed metric, PASR, which regards an attack as successful only when at least one pedestrian is completely undetected. Simpler loss forms (\eg, obj, cls, or their product) fail to explicitly minimize all three factors and may leave partial detections that still result in PASR failure. To sum up, this section offers theoretical support for the superior transferability of our attack in terms of PASR across unseen detectors and datasets, as consistently validated by our experiments.

\end{document}

%% file: _header.tex
\usepackage{graphicx}
\usepackage[table]{xcolor}
\usepackage[normalem]{ulem}
\useunder{\uline}{\ul}{}
\usepackage{tikz}
\usepackage{amsfonts}
\usepackage{xspace} 
\usepackage{amsmath}
\usepackage{mathrsfs}
\usepackage{amssymb}
\usepackage{enumitem}
\usepackage{amsmath}
\usepackage{amsthm}
\usepackage{multirow}
\usepackage{makecell}
\usepackage{caption}
\usepackage{subcaption}
\usepackage{float}
\usepackage{booktabs}
\usepackage{balance}
\usepackage{tcolorbox}
\usepackage{xurl}
\usepackage{hyperref}
\usepackage{tabularx}
\usepackage{cleveref}

\usepackage{cite}

\usepackage{diagbox}
\usepackage{placeins}
\setlist[itemize]{leftmargin=3mm}
\usepackage{dsfont}
\usepackage[ruled,vlined,linesnumbered]{algorithm2e}
\usepackage{algorithm2e}
\usepackage{algorithmicx}



\usepackage{siunitx}
\def\eg{\emph{e.g.,}\xspace}

\def\ie{\emph{i.e.,}\xspace}